\documentclass[final,sort&compress,a4paper,twocolumn]{article}
\usepackage{arxiv} 
\usepackage{authblk}
\usepackage[utf8]{inputenc}
\usepackage{amsmath}
\usepackage{amssymb}
\usepackage{graphicx}
\usepackage{acro}
\usepackage[hyphens]{url}
\usepackage[hidelinks]{hyperref}
\usepackage[capitalize]{cleveref}
\usepackage{enumitem}
\usepackage{multirow}
\usepackage{multicol}
\usepackage{siunitx}
\usepackage{float}
\usepackage[flushleft]{threeparttable}
\usepackage[table,xcdraw]{xcolor}
\usepackage[switch]{lineno}
\usepackage{array}
\usepackage{soul}
\usepackage{parskip}
\usepackage{caption}
\usepackage{subcaption}
\usepackage{tabularx}
\usepackage{placeins}
\usepackage{makecell}
\usepackage{tikz}
\usepackage[export]{adjustbox}
\usepackage{framed}
\usepackage[labelfont=bf]{caption}
\usepackage[font={small}]{caption}
\newcolumntype{L}[1]{>{\hsize=#1\hsize\raggedright\arraybackslash}X}%
\newcolumntype{R}[1]{>{\hsize=#1\hsize\raggedleft\arraybackslash}X}%
\newcolumntype{C}[1]{>{\hsize=#1\hsize\centering\arraybackslash}X}%
\usepackage{setspace}
\begin{document}

\title{An automated, geometry-based method for hippocampal shape and thickness analysis} 

\author[1]{Kersten Diers}
\author[2]{Hannah Baumeister}
\author[3,4,5]{Frank Jessen}
\author[6,7,8]{Emrah Düzel}
\author[2,9]{David Berron}
\author[1,10,11]{Martin Reuter}

\affil[1]{\small{AI in Medical Imaging, German Center for Neurodegenerative Diseases (DZNE), Bonn, Germany}}
\affil[2]{\small{Clinical Cognitive Neuroscience Group, German Center for Neurodegenerative Diseases (DZNE), Magdeburg, Germany}}
\affil[3]{\small{Clinical Alzheimer’s Disease Research, German Center for Neurodegenerative Diseases (DZNE), Bonn, Germany}}
\affil[4]{\small{Department of Psychiatry, Medical Faculty, University of Cologne, Cologne, Germany}}
\affil[5]{\small{Excellence Cluster on Cellular Stress Responses in Aging-Associated Diseases, University of Cologne, Cologne, Germany}}
\affil[6]{\small{Clinical Neurophysiology and Memory Group, German Center for Neurodegenerative Diseases (DZNE), Magdeburg, Germany}}
\affil[7]{\small{Institute of Cognitive Neurology and Dementia Research, Otto-von-Guericke University, Magdeburg, Germany}}
\affil[8]{\small{Institute of Cognitive Neuroscience, University College London, London, United Kingdom}}
\affil[9]{\small{Clinical Memory Research Unit, Department of Clinical Sciences Malmö, Lund University, Lund, Sweden}}
\affil[10]{\small{A.A. Martinos Center for Biomedical Imaging, Massachusetts General Hospital, Boston MA, USA }}
\affil[11]{\small{Department of Radiology, Harvard Medical School, Boston MA, USA}}

% --------------------------------------------------------------

\date{}

\twocolumn[
\begin{@twocolumnfalse}
  \maketitle
  \vspace{-5ex}
  \begin{abstract}

The hippocampus is one of the most studied neuroanatomical structures due to its involvement in attention, learning, and memory as well as its atrophy in ageing, neurological, and psychiatric diseases. Hippocampal shape changes, however, are complex and cannot be fully characterized by a single summary metric such as hippocampal volume as determined from MR images. In this work, we propose an automated, geometry-based approach for the unfolding, point-wise correspondence, and local analysis of hippocampal shape features such as thickness and curvature. 
Starting from an automated segmentation of hippocampal subfields, we create a 3D tetrahedral mesh model as well as a 3D intrinsic coordinate system of the hippocampal body. From this coordinate system, we derive local curvature and thickness estimates as well as a 2D sheet for hippocampal unfolding.
We evaluate the performance of our algorithm with a series of experiments to quantify neurodegenerative changes in Mild Cognitive Impairment and Alzheimer's disease dementia. We find that hippocampal thickness estimates detect known differences between clinical groups and can determine the location of these effects on the hippocampal sheet. Further, thickness estimates improve classification of clinical groups and cognitively unimpaired controls when added as an additional predictor. Comparable results are obtained with different datasets and segmentation algorithms.
Taken together, we replicate canonical findings on hippocampal volume/shape changes in dementia, extend them by gaining insight into their spatial localization on the hippocampal sheet, and provide additional, complementary information beyond traditional measures. 
We provide a new set of sensitive processing and analysis tools for the analysis of hippocampal geometry that allows comparisons across studies without relying on image registration or requiring manual intervention.

\end{abstract}

\vspace{4ex}


\vspace{2ex}

\noindent \textbf{\textit{Corresponding author}}: Martin Reuter (\texttt{martin.reuter [at] dzne.de})

\end{@twocolumnfalse}
]

\section{Introduction}

%Relevance 

The hippocampus, one of the phylogenetically oldest structures of the human brain, supports fundamental cognitive processes such as attention \cite{goldfarb2016memory}, learning, and memory \cite{gabrieli1997separate, maguire2001neuroimaging, henke1997human}. Hippocampal damage and dysfunction have been associated with important neurological and psychiatric conditions such as Alzheimer's disease (AD) \cite{adler2018characterizing,fjell2014normal}, epilepsy \cite{cendes1993mri, jack1994mri}, or major depression \cite{videbech2004hippocampal}. Beyond basic research, hippocampal volume has been proposed as a diagnostic biomarker to detect early disease changes \cite{jack2018nia, frisoni2017strategic} and as an outcome marker for monitoring therapeutic efficacy in clinical trials \cite{cummings2019role, mattsson2015revolutionizing}. 

%Hippocampal anatomy

The hippocampal formation is embedded into the medial temporal lobe \cite{amaral1990}, where it connects to adjacent cortical areas such as the entorhinal, perirhinal, and parahippocampal cortices \cite{insausti1998mr}. The hippocampus itself is not a homogeneous structure, but consists of anatomically and histologically distinct substructures, including the subiculum, the four cornu ammonis (CA) subfields, and the dentate gyrus \cite{duvernoy1998}. While the exact borders and subdivisions are still a topic of debate, its macroscopic morphology can be described as resembling a spiral or Swiss roll, where the subiculum and CA subfields 1 to 3 form an outer layer, wrapping the interior CA4 and dentate gyrus. The border between these layers is indicated by the stratum radiatum, lacunosum, moleculare (SRLM), which is easily identified on T2-weighted MR images and often used as an anatomical landmark in hippocampal segmentation. Along the anterior-posterior or longitudinal axis, a division can be made into the head, body, and tail, with folding complexity increasing from tail towards the head. Recent evidence points to a functional dissociation not only for the different subfields, but also along the long axis \cite{genon2021many}. 

%Analysis approaches for quantification 

Structural MRI is the primary means for in-vivo anatomical analysis of the hippocampus in humans, and advances in MR technology, in particular the development of dedicated imaging sequences and the acquisition at higher field strengths, have led to increased spatial resolution and contrast in MR images, which makes the substructures of the hippocampus accessible for human neuroimaging \cite{berron2017protocol}. In contrast to the advancement in measurement methods, analysis methods have seen less progress to date, even though more sensitive tools to quantify (local) hippocampal changes promise to identify disease or therapeutic effects at an earlier stage or with smaller sample sizes. Traditional analysis approaches create a manual or automated segmentation of the hippocampus and/or its subfields, and quantify the corresponding regional volumes by means of voxel-counting \cite{vanleemput2009automated, yushkevich2009high}. While pragmatic and straightforward, these measures are susceptible to partial-volume effects and registration inaccuracies, and cannot detect subtle feature changes that do not result in volume changes. Further, in situations where there is insufficient intensity information / contrast within the image, the segmentations and specifically the boundaries of the subfields will, to a large extent, be driven by the deformation of an atlas template rather than actual intensity differences, and will, therefore, be less reliable. Finally, these methods provide volume estimates per individual subfield and, therefore, still give high-level summary measures instead of true point-wise measurements. Geometrical shape models of the hippocampus on the other hand may provide complementary information beyond voxel-based methods, but until now typically focus on the hippocampus as a whole, and not its substructures \cite{achterberg2014hippocampal}.

%Related work

Recognizing the need for a more fine-grained analysis, Zeineh et al.\ \cite{zeineh2000, ekstrom2009} proposed a procedure to unfold hippocampal anatomy and localize functional activation onto a two-dimensional, flat map of the hippocampus, similar to retinotopic imaging and analysis. It is based on a manual segmentation of gray and white matter and cerebrospinal fluid in the medial temporal lobe as well as manual demarcations of the boundaries between the subregions of the hippocampus. The gray matter segmentation is split into layers, each layer is stretched to a planar representation, and is subsequently recombined with the other layers to give a 2D flat map of the hippocampus. Boundaries between hippocampal regions are then projected onto this map. A common template is created from averaged individual maps, including the individual subfield boundaries, to account for individual differences in anatomy. Individual maps as well as coregistered functional data are then transformed to match the template by means of a nonlinear warping algorithm. Applications of this technique have resulted, for example, in insights about the role of hippocampal subfields in encoding vs.\ retrieval processes in human memory \cite{zeineh2003dynamics} as well as associations of hippocampal thinning with risk factors for Alzheimer's disease \cite{donix2013cardiovascular}.

The ASHS algorithm \cite{yushkevich2015} focuses on localized thickness estimation and performing group analyses in a common template space. The algorithm works by warping a template surface of the whole hippocampus into the space of individual segmentations (using the diffeomorphic deformation fields generated during the construction of the template). Hippocampal thickness is then computed for each surface point by extracting the Voronoi skeleton of the surface, removing any branches from the skeleton, and computing the distance from each point on the surface to the closest point on the pruned skeleton. This allows for an analysis of point-wise thickness, although not on an unfolded surface, but on the surface of the population template that resembles the anatomy of the whole hippocampus.

Recent work on hippocampal unfolding has combined geometrical modeling with histological validation. DeKraker et al.\ \cite{dekraker2018} manually traced the hippocampal sulcus and the stratum radiatum, lacunosum, moleculare, by which much of the morphology of the hippocampus is captured, and then conducted a semi-automated segmentation of hippocampal grey matter. Boundaries for three dimensions across the hippocampus were defined at the borders between the cortex and the subiculum as well as between the dentate gyrus and the SRLM for the proximal-distal axis, at the anterior and posterior ends of the hippocampus for the longitudinal axis, and at the outer hippocampal surface and the SRLM for the laminar axis. Laplace’s equation was then solved in voxel space to determine potential fields along each axis. Based on these fields, coordinates in 2D space (long axis and proximal-distal axis) were derived to create an unfolded representation of the hippocampus for the mapping of hippocampal thickness, myelin content, and subfield labels. This work has recently been extended to identify the boundaries of hippocampal subfields in an automated fashion by using morphological features (such as thickness, curvature, gyrification) and also laminar features (based on histology) of the unfolded hippocampus \cite{dekraker2020hippocampal}. In its most recent version \cite{dekraker2022automated}, the algorithm also incorporates automated segmentation by means of an nnU-Net \cite{isensee2021nnu} that was trained with manually-corrected hippocampal segmentations from the HCP-1200 Young Adult dataset \cite{van2013wu} and tested on a wider range of datasets. Further geometry-based analyses using Laplacian methods and tetrahedral meshes have been performed in the cortex \cite{wang2017towards, fan2021tetrahedral}, but not in the hippocampus until now.

%Our approach

In summary, previous work has resulted in approaches for the unfolding of the hippocampus, the imposition of a coordinate system, the creation of a common anatomical space, and for thickness estimation and the mapping of other signals to the unfolded hippocampus. Here, we present an algorithm that builds upon and extends these previous developments. Our method estimates local hippocampal thickness and additional morphological features by means of a sheet representation and an intrinsic coordinate system of the hippocampal body. This corresponds to an unfolding of the hippocampus and simultaneous creation of a reference frame that is consistent across individual hippocampal geometries. 

The primary innovation of our approach is the application of differential geometry operators in a flexible mesh model of the hippocampus. This means that after segmentation all further operations are performed with a triangle (boundary) and tetrahedral mesh, which represents an anatomically adaptive discretization that is no longer dependent on a rigid voxel grid. As a consequence, these models are not constrained to the voxel grid any longer, and can hence more accurately capture the intricate geometry of the hippocampal anatomy.

A second innovation is the application of a curvature-aware, anisotropic Laplace operator, which provides the automated detection of landmarks on the hippocampal mesh. This obviates the need for manual delineation of these features. In addition, curvature is also used to align hippocampal thickness estimates along the medial/lateral axis, since hippocampal size changes along this dimension can impact correspondence of the coordinate grid across clinical groups, and may confound thickness estimates if not accounted for.

Finally, our algorithm requires no manual intervention, has a relatively short runtime ($<25$~mins per hemisphere), is designed to work with different automated segmentation algorithms, and is exhaustively tested and validated in a range of prototypical application scenarios. Specifically, we conduct an evaluation of our algorithm in two large, independent samples, additionally investigating the impact of two common automated hippocampal segmentation algorithms.

We expect that the core features of this approach, the point-wise correspondence across the hippocampal sheets (across time, individuals or hemispheres), permit a more precise characterization of changes in hippocampal size and shape compared to traditional voxel- or atlas-based summary measures. Since our algorithm does not depend on potentially unreliable boundaries between hippocampal subfields, it permits the creation of custom regions of interest as well as the localization of effects that extend across subfields. Finally, the analysis of geometric features in addition to thickness measures is expected to open up new avenues for characterizing hippocampal shape changes in health and disease as well as across time.

\section{Methods}

In this section, we give a technical description of the proposed algorithm and its incorporation into a hippocampal shape and thickness analysis pipeline, followed by an overview of the evaluation framework and empirical analyses. The hippocampal shape and thickness analysis (HIPSTA) packaget will be available at \url{https://github.com/Deep-MI/Hipsta} upon publication.

\subsection{Methodology of the hippocampal shape and thickness analysis}

The algorithm builds upon existing hippocampal segmentations on high-resolution T2 MR images, i.e.\ images with approximately 0.5 mm in-plane resolution and approximately 1.5-2.5 mm slice thickness. The segmentations can be created in an automated or a manual fashion. In particular, the method is applicable to both the output of FreeSurfer's hippocampal subfields segmentation \cite{iglesias2015} as well as the hippocampal analysis pipeline in ASHS \cite{yushkevich2015}. In addition, manual labeling protocols \cite{yushkevich2015quantitative, wisse2017harmonized} can also provide suitable inputs as long as they provide labels for the subiculum and the CA substructures of the hippocampus.

\subsubsection*{Step 1: Image segmentation and shape definition}

\begin{figure}[!hbt]
    \centering
    \begin{framed}
    \includegraphics[width=\textwidth]{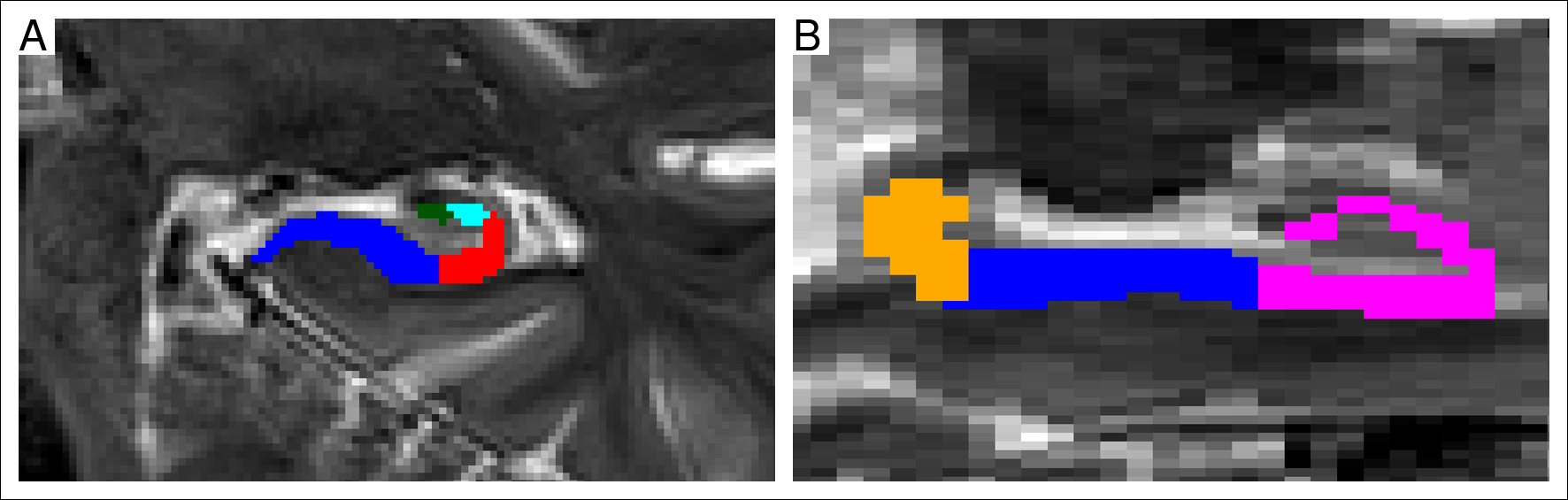}
    \end{framed}
    \captionsetup{width=0.475\textwidth}
    \caption{Cut-out of an MR image of the hippocampus, with ASHS-derived labels for the subfields used for unfolding: (a) subiculum (blue), CA1 (red), CA2 (cyan), CA3 (green), (b) same structures as in (a), plus labels for the head (magenta) and tail (orange).}
    \label{fig:m-1}
\end{figure}

The algorithm employs and combines labels of the presubiculum, subiculum, CA1, CA2, and CA3 subfields (Figure~\ref{fig:m-1}(a)). Two or more of these regions may share the same label; for example, the FreeSurfer segmentation does not distinguish between CA2 and CA3, and the ASHS segmentation has a single label for the presubiculum and subiculum. CA4 and the dentate gyrus are not included as they represent anatomically distinct structures. If a separate label for the molecular layer, i.e.\ the most superficial layer of the hippocampus proper, is present, all voxels of this structure are assigned to the nearest subfield of the hippocampal body. Further, if labels for the hippocampal head and tail exist (such as in Figure~\ref{fig:m-1}(b)), these will be used to define the anterior and posterior extent of the region considered for unfolding; otherwise, the algorithm uses the most anterior slice with either the CA2 or CA3 label and the most posterior slice with CA1 and CA2/CA3 labels to define the anterior and posterior extent, and restricts the segmentation accordingly. This is a technical restriction, since -- at the current (thick-slice) voxel resolutions -- the tail is lacking sufficient anatomical detail and the head folds sideways onto itself preventing a reliable unfolding. Prior to surface extraction, small holes or protrusions in the binarized segmentation are corrected by means of a repeated closure operation (dilation and erosion). 

\subsubsection*{Step 2: Mesh generation}

\begin{figure}[!hbt]
    \centering
    \begin{framed}
        \includegraphics[width=\textwidth]{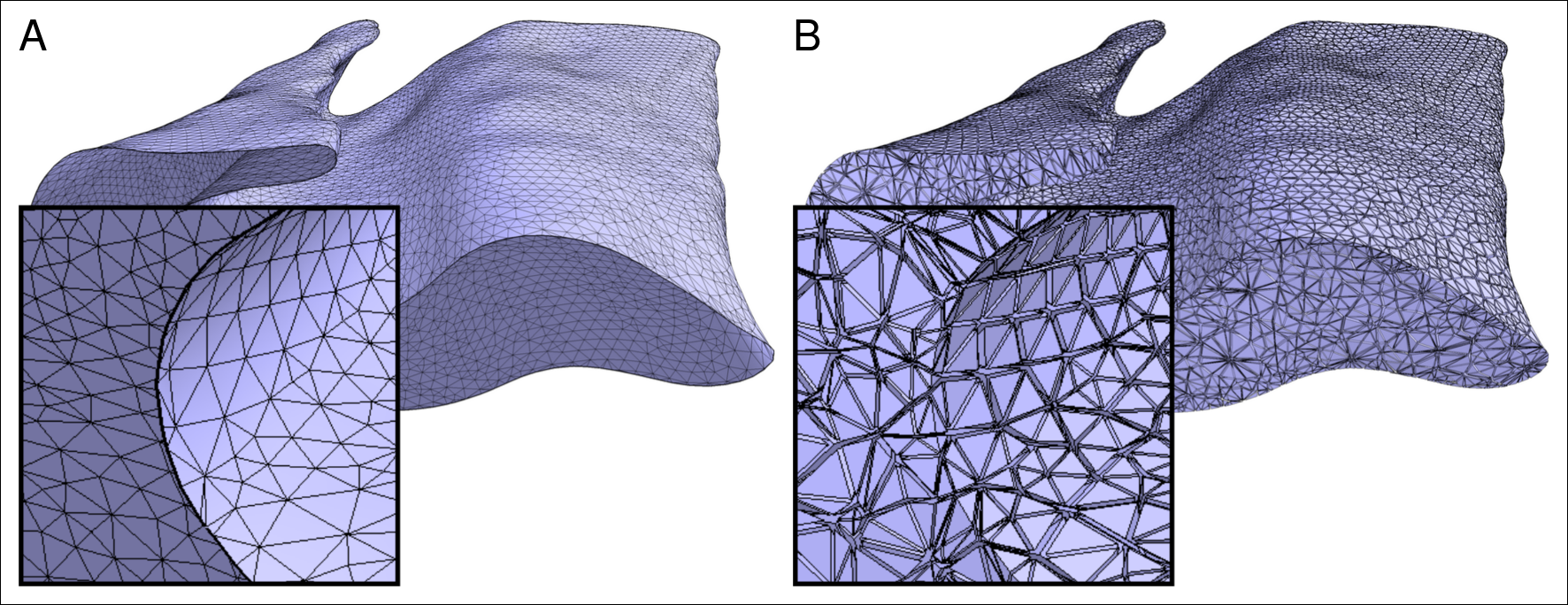}
    \end{framed}
    \captionsetup{width=0.475\textwidth}
    \caption{Mesh representations of the hippocampal body, using (a) 2D surface triangles and (b) 3D tetrahedral volume elements.}
    \label{fig:m-2}
\end{figure}

An initial surface representation of the hippocampal body is obtained via the marching cube algorithm \cite{lorensen1987}, which provides a mesh of triangles, represented by edges and vertices (3D point coordinates, Figure~\ref{fig:m-2}(a)). Mesh quality is further improved by mild mesh smoothing. The resulting surface mesh is a closed 2D manifold embedded in 3D space (i.e.\ a boundary representation) and has no representation of the interior of the hippocampal body. We, therefore, create a 3D tetrahedral mesh model of the full hippocampal body using the \textit{GMSH} software package \cite{geuzaine2009}, filling the shell's interior completely with tetrahedral volumetric elements (Figure~\ref{fig:m-2}(b)). We also transfer the labels of the hippocampal subfields from the voxel-based to the vertex-based representations using nearest neighbor mapping and additionally create labels for the boundaries of the hippocampal body with the head and tail.

\subsubsection*{Step 3: Identification of boundaries}

\begin{figure}[!hbt]
    \begin{framed}
        \centering
    \includegraphics[width=\textwidth]{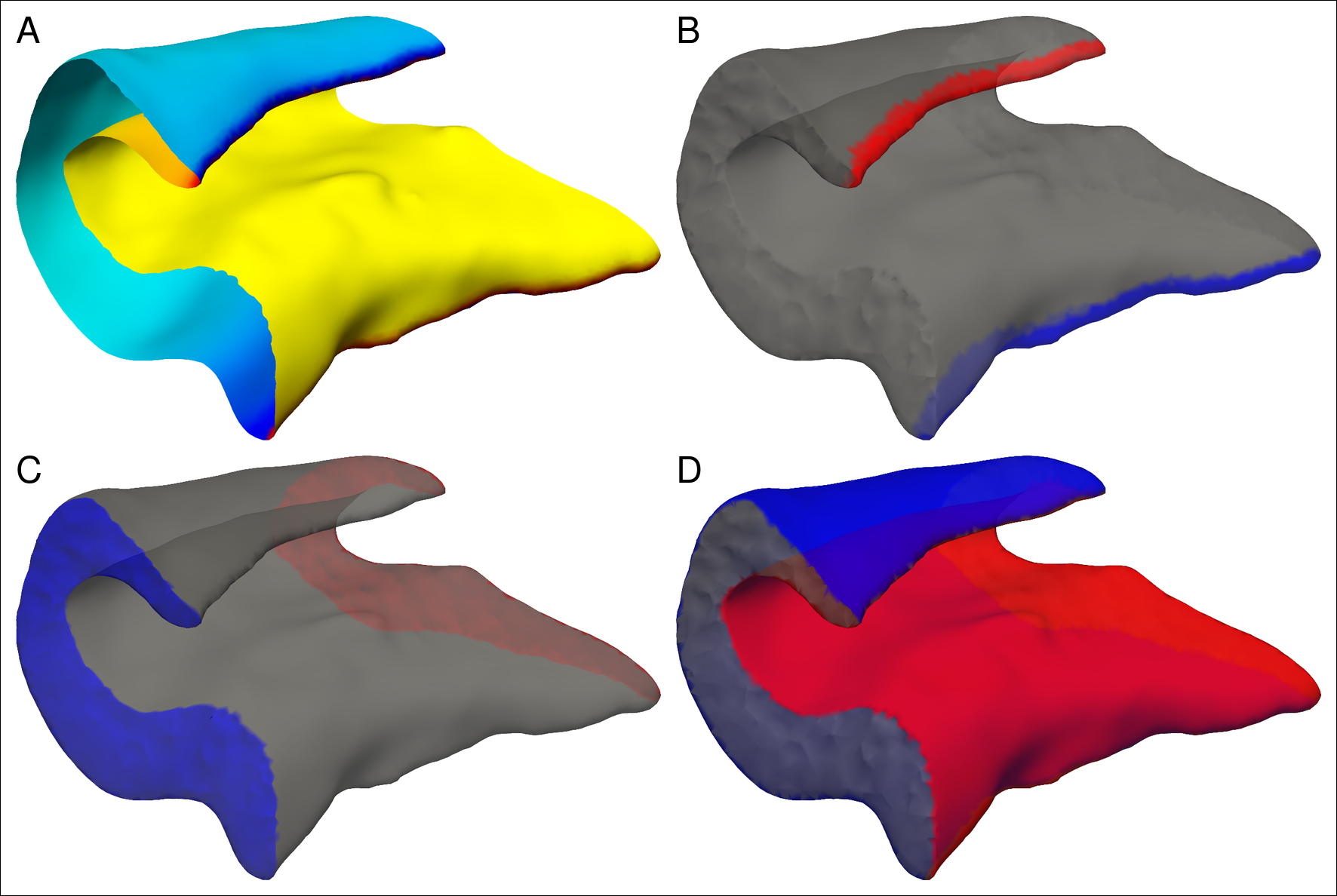}
    \end{framed}
    \captionsetup{width=0.475\textwidth}
    \caption{Identification of boundaries on the hippocampal mesh: (a) first eigenvalue of the anisotropic Laplace-Beltrami operator, (b) medial/lateral boundaries, (c) anterior/posterior boundaries, (d) interior/exterior boundaries.}
    \label{fig:m-3}
\end{figure}

We next identify boundaries on the hippocampal mesh, which are needed to establish parameter functions across the mesh. A total of three parameter functions are estimated: between the lateral (distal) and medial (proximal), the anterior and posterior, and the interior/exterior boundaries of the hippocampus (Figure~\ref{fig:m-3}). Due to the existing segmentation labels, we already know anterior and posterior boundaries at the transition of hippocampal body and tail as well as body and head (Figure~\ref{fig:m-3}(c)). The medial (presubiculum/entorhinal cortex) and lateral (CA3/CA4) boundary curves are located at the proximal and distal high curvature regions of the hippocampus and can be smoothly estimated via the zero level sets of an appropriate anisotropic surface Laplace-Beltrami eigenfunction (see Appendix~\ref{sec:albo} for a detailed description of this procedure). For this purpose, both the 2D and 3D meshes are cut open at the transitions towards the hippocampal head and tail. On the resulting open cylinder-like surface, we compute the first eigenfunction of the anisotropic (i.e.\ curvature-aware) Laplace-Beltrami operator \cite{andreux2014}, with Neumann boundary conditions (Figure~\ref{fig:m-3}(a)). Anisotropy parameters are chosen such that the zero level sets of the first eigenfunction are attracted to the high curvature zones of the hippocampal surface. These are precisely the medial boundary curve $C_m$ between the presubiculum and the adjacent entorhinal cortex and the lateral boundary $C_l$ between CA3 and CA4, which in turn allow to define the medial/lateral (Figure~\ref{fig:m-3}(b)) and interior/exterior parts (Figure~\ref{fig:m-3}(d)) of the hippocampal body.

\subsubsection*{Step 4: Mesh parametrization} 

\begin{figure}[!hbt]
    \begin{framed}
        \centering
        \includegraphics[width=\textwidth]{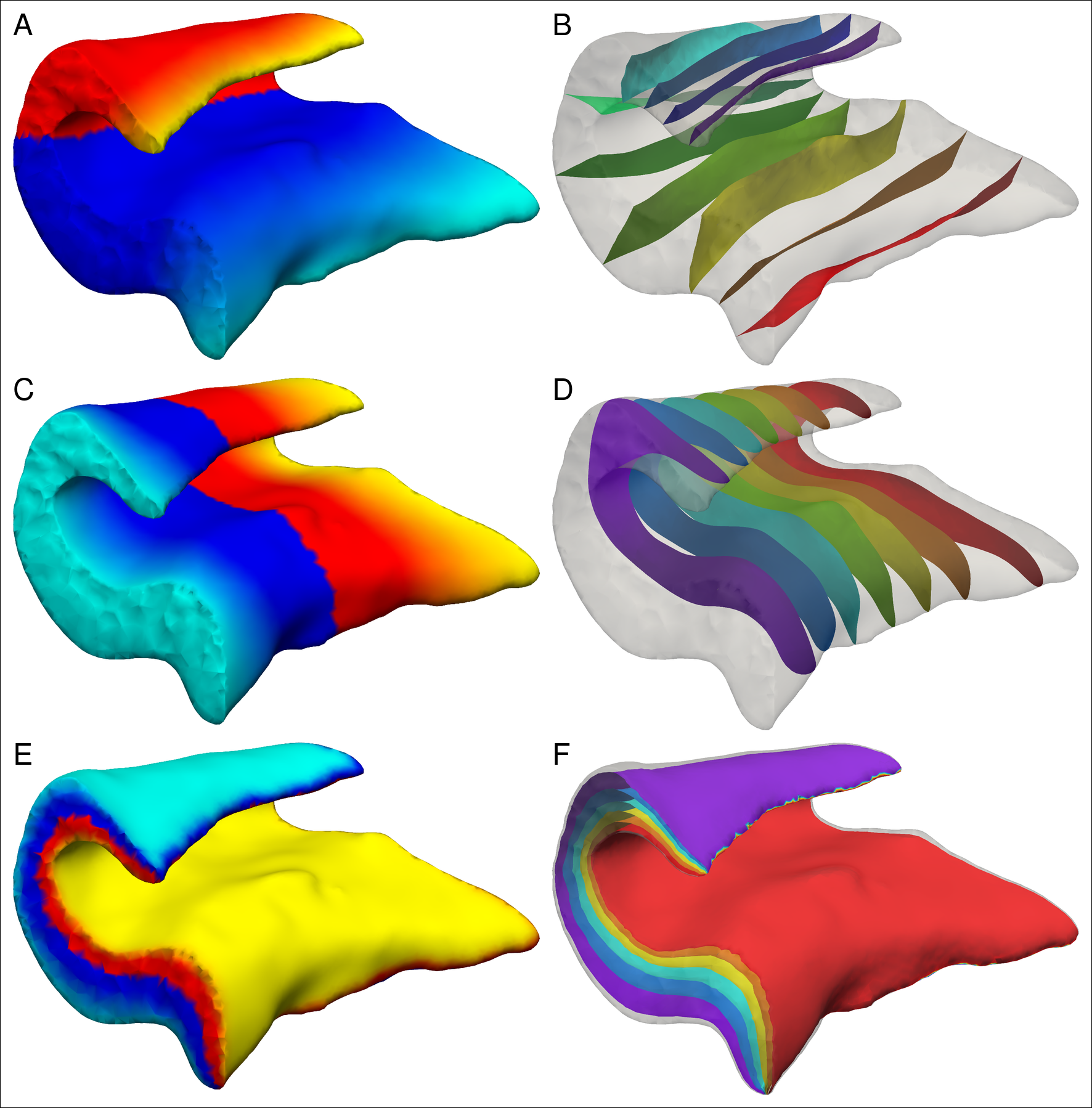}
    \end{framed}
    \captionsetup{width=0.475\textwidth}
    \caption{Mesh parametrization by solving Laplace equations in the (a) medial/lateral, (c) anterior/posterior, and (e) interior/exterior directions. Subfigures (b), (d), (f) show the corresponding levelsets of these functions.}
    \label{fig:m-4}
\end{figure}

Given the boundaries, we can now find a harmonic map to a (degenerated) unit cube. This mapping is bijective everywhere, except at the two boundary curves $C_l$ and $C_m$ (as the cube is convex, Rado-Kneser-Choquet Theorem). The mapping can be computed by solving three Laplace equations (via FEM) with the corresponding boundary conditions (0~and~1 Dirichlet conditions at the opposing boundary surfaces or edges and Neumann conditions elsewhere). Figures~\ref{fig:m-4}(a), (c), and (e) show parametrizations of the tetrahedral mesh by the three Laplace functions that run into medial/lateral, anterior/posterior, and interior/exterior directions. Importantly, these functions are defined not only at the surface, but also in the volume of the mesh, as indicated by the level-sets of the functions (Figures~\ref{fig:m-4}(b), (d), and (f)). Jointly, these three functions define an intrinsic coordinate system for the hippocampal body.

\subsubsection*{Step 5: Grid and thickness estimation}

\begin{figure}[!hbt]
    \begin{framed}
        \centering
        \includegraphics[width=\textwidth]{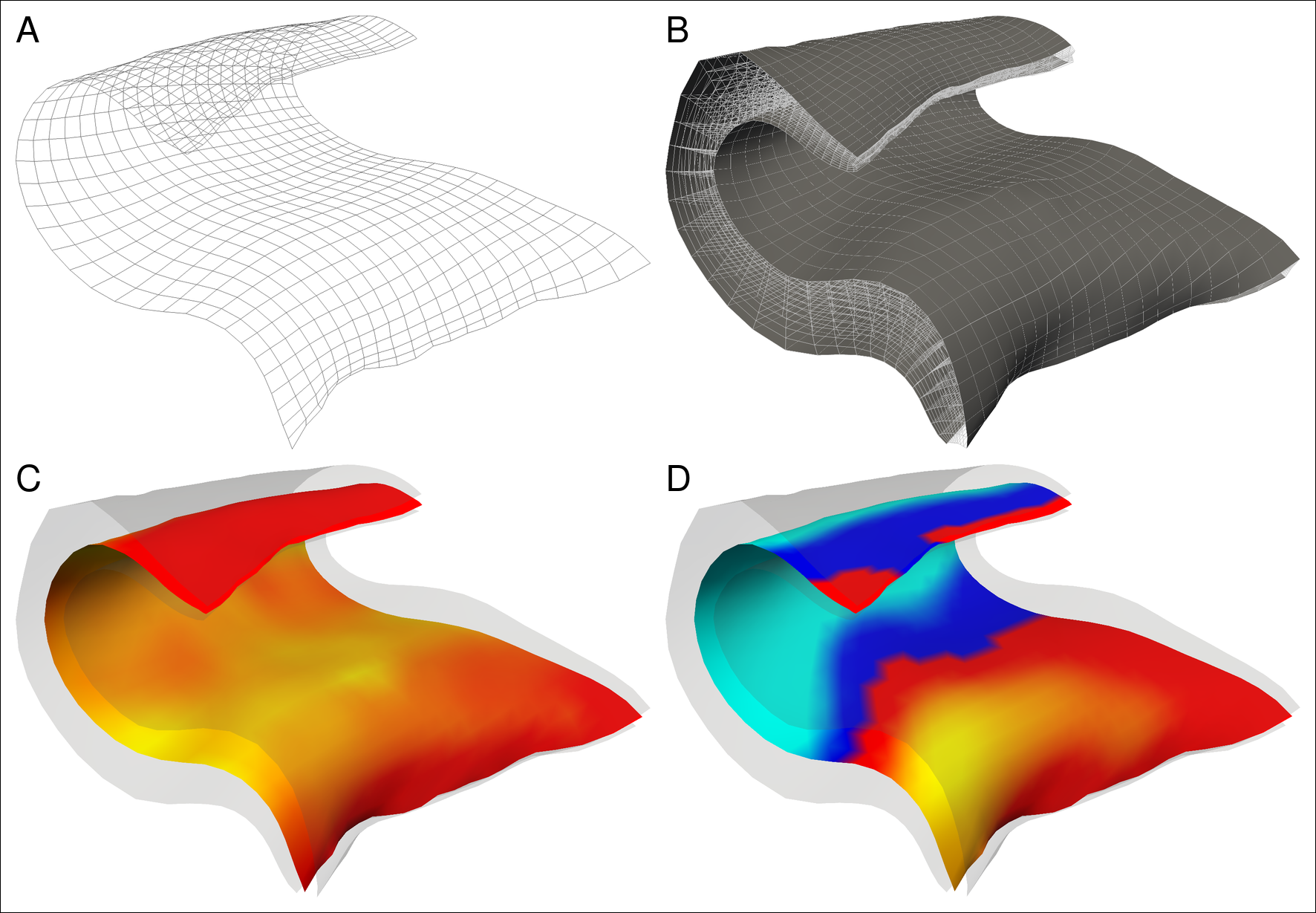}
    \end{framed}
    \captionsetup{width=0.475\textwidth}
    \caption{Maps and shape characteristics derived from the mesh parametrization: (a) 2D coordinate system on the mid-surface, (b) 3D coordinate system and streamlines, (c) thickness estimates overlaid onto the mid-surface (yellow/red colors indicate higher/lower thickness), (d) curvature of the mid-surface (cyan/blue colors indicate higher/lower negative curvature, yellow/red colors indicate higher/lower positive curvature).}
    \label{fig:m-5}
\end{figure}

An approximate mid-surface of the hippocampal body can be obtained as the $0.5$ level set in the interior/exterior direction, onto which we prescribe a regular $N \times M$ grid in the remaining two directions (anterior/posterior and medial/lateral; Figure~\ref{fig:m-5}(a) and \ref{fig:m-5}(b)). Hippocampal thickness is then computed by calculating distances along the interior/exterior streamlines at each grid point. The mid-surface grid serves as a reference frame that is anatomically consistent across individual hippocampi, since it is defined by their intrinsic geometries. By following the streamlines, the mid-surface grid can also be carried to the interior or to the exterior boundaries, giving a 3D grid. These grids can be used for point-wise statistical comparison and visualization of features such as thickness (Figure~\ref{fig:m-5}(c)), curvature (Figure~\ref{fig:m-5}(d)), and for the projection of subfield labels or other volumetric data onto the hippocampal sheet (mid-surface) or its outer surface.

\subsubsection*{Step 6: Alignment and statistics}

After the determination of individual hippocampal thickness, we prepare the data for statistical analysis with an additional postprocessing step, a curvature-based alignment procedure. This step is motivated by the notion that anatomical changes in the hippocampus in aging or disease may not only encompass a reduction in thickness, i.e.\ in the interior/exterior dimension, but also shrinkage in the medial/lateral or the anterior/posterior directions.
For this reason, we employ a curvature-based spatial alignment procedure to correct for potential shifts of the coordinate system. Assuming that the overall shape of the hippocampus -- as characterized by its curvature -- remains intact, individual curvature estimates (averaged across the anterior/posterior dimension) are aligned by means of an interpolation procedure so that their maxima and minima are located at the same locations along the medial/lateral axis \cite{Wrobel2018}; see Appendix \ref{sec:curv-align} for details). The resulting shifting parameters are then applied to the individual thickness estimates. As a result of this procedure, the localized thickness estimates are comparable across individuals and groups, which is a prerequisite for subsequent point-wise statistical analysis. A comparison of the evaluation of aligned vs.\ non-aligned thickness estimates is provided in Section \ref{sec:align} in the Appendix.
Figure~\ref{fig:m-6} shows curvature and thickness profiles across the medial/lateral axis of the hippocampus. These profiles give a concise representation of where differences in curvature and thickness are present, in particular since effects do not vary much along the longitudinal axis. Panels (a) and (c) show the original, non-aligned curvature and thickness, respectively, and panels (b) and (d) show the curvature and thickness after alignment. The overall pattern is similar, but the registered data are more aligned across groups. 

\begin{figure*}[!hbt]
    \centering
    \begin{minipage}[!hbt]{0.75\linewidth}
        \begin{framed}
        \centering
        \includegraphics[width=\textwidth]{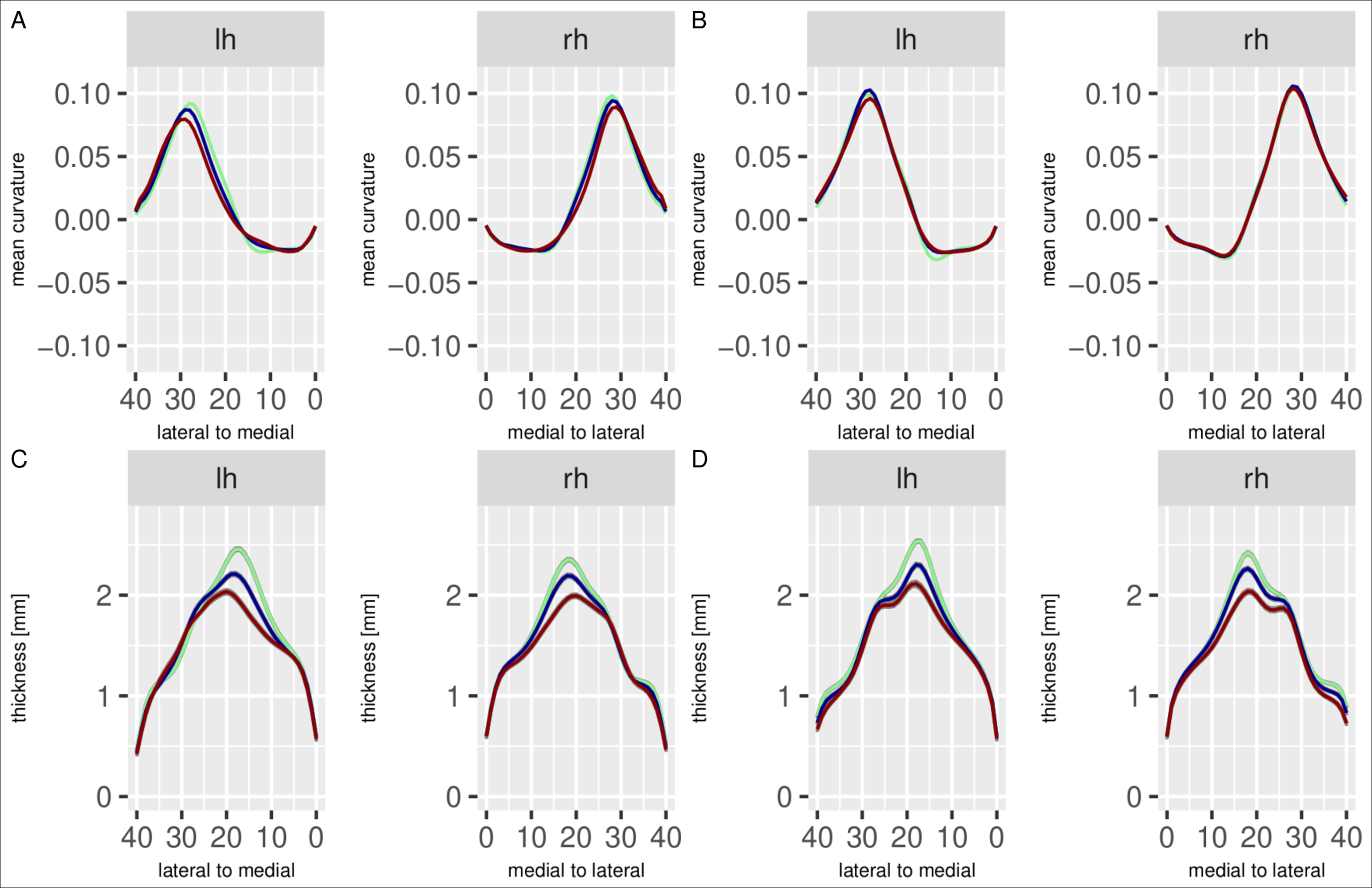}
        \end{framed}
    \end{minipage}
    \captionsetup{width=0.75\textwidth}
    \caption{Curvature and thickness profiles in the left (lh) and right (rh) hemisphere, averaged across the medial/lateral axis of the hippocampus. Colors indicate diagnostic groups, taken from the DELCODE / ASHS dataset: cognitively unimpaired controls (green), mild cognitive impairment (blue), dementia (red). Panels show (a) original curvature, (b) aligned curvature, (c) original thickness, (d) aligned thickness.}
    \label{fig:m-6}
\end{figure*}

\subsection{Morphometric features of the hippocampal shape and thickness analysis}

\begin{figure}[!hbt]
    \begin{framed}
        \centering
        \includegraphics[width=\textwidth]{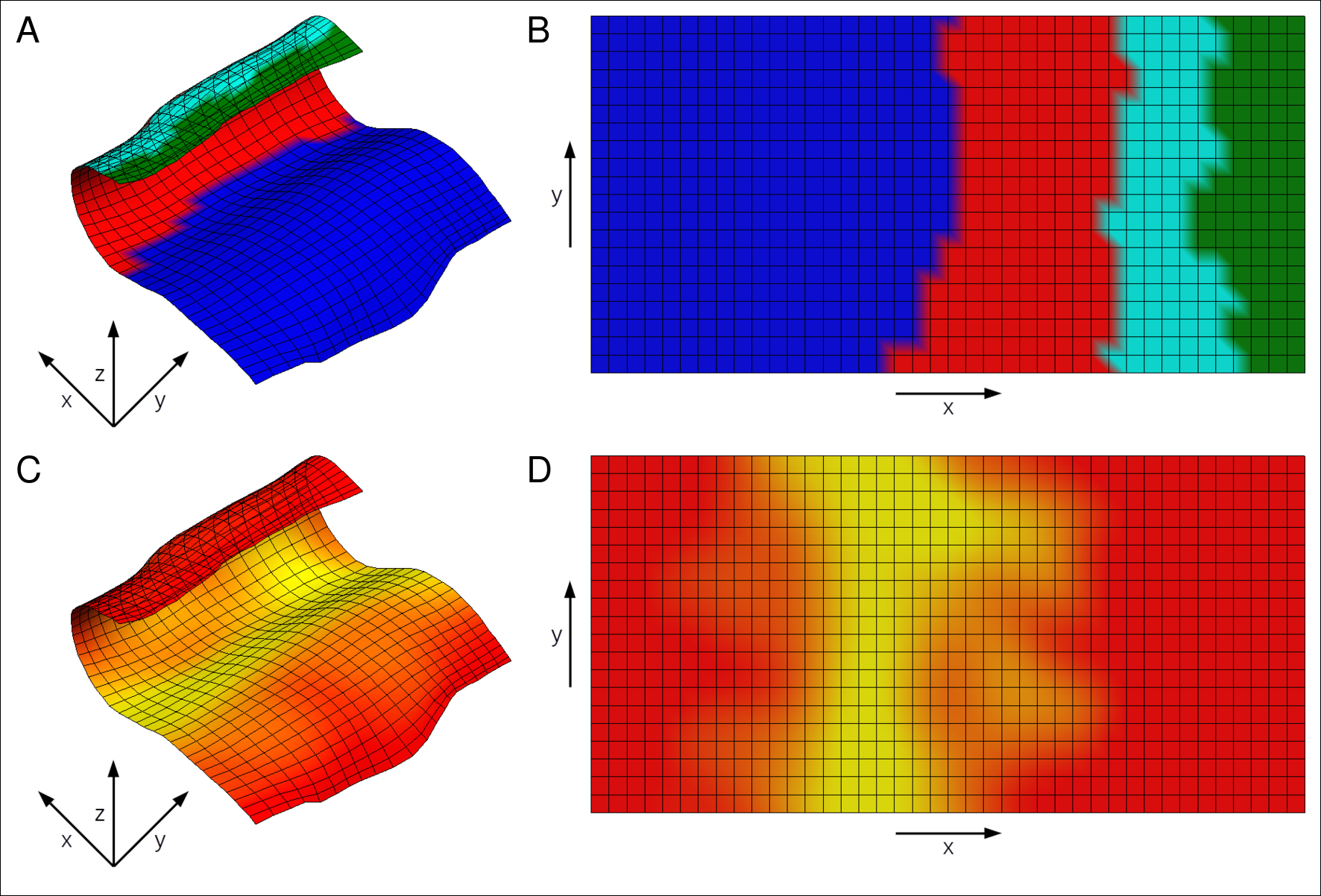}
    \end{framed}
    \captionsetup{width=0.475\textwidth}
    \caption{Correspondence between the mid-surface of a left hippocampus and a 2D flat map; the $x$ axis follows the hippocampal sheet from medial (subiculum) to lateral (CA3), the $y$ axis runs from posterior to anterior, and the $z$ axis is always perpendicular to the hippocampal sheet, pointing towards superior in medial regions (such as the subiculum) and towards inferior in lateral regions (such as CA3). Panels (a) and (b) show labels for the subiculum (blue), CA1 (red), CA2 (cyan), and CA3 (green). Panels (c) and (d) show thickness estimates both for the mid-surface and the 2D grid, with thickness increasing from red to yellow.}
    \label{fig:m-7}
\end{figure}

\begin{figure}[!hbt]
    \begin{framed}
        \centering
        \includegraphics[width=\textwidth]{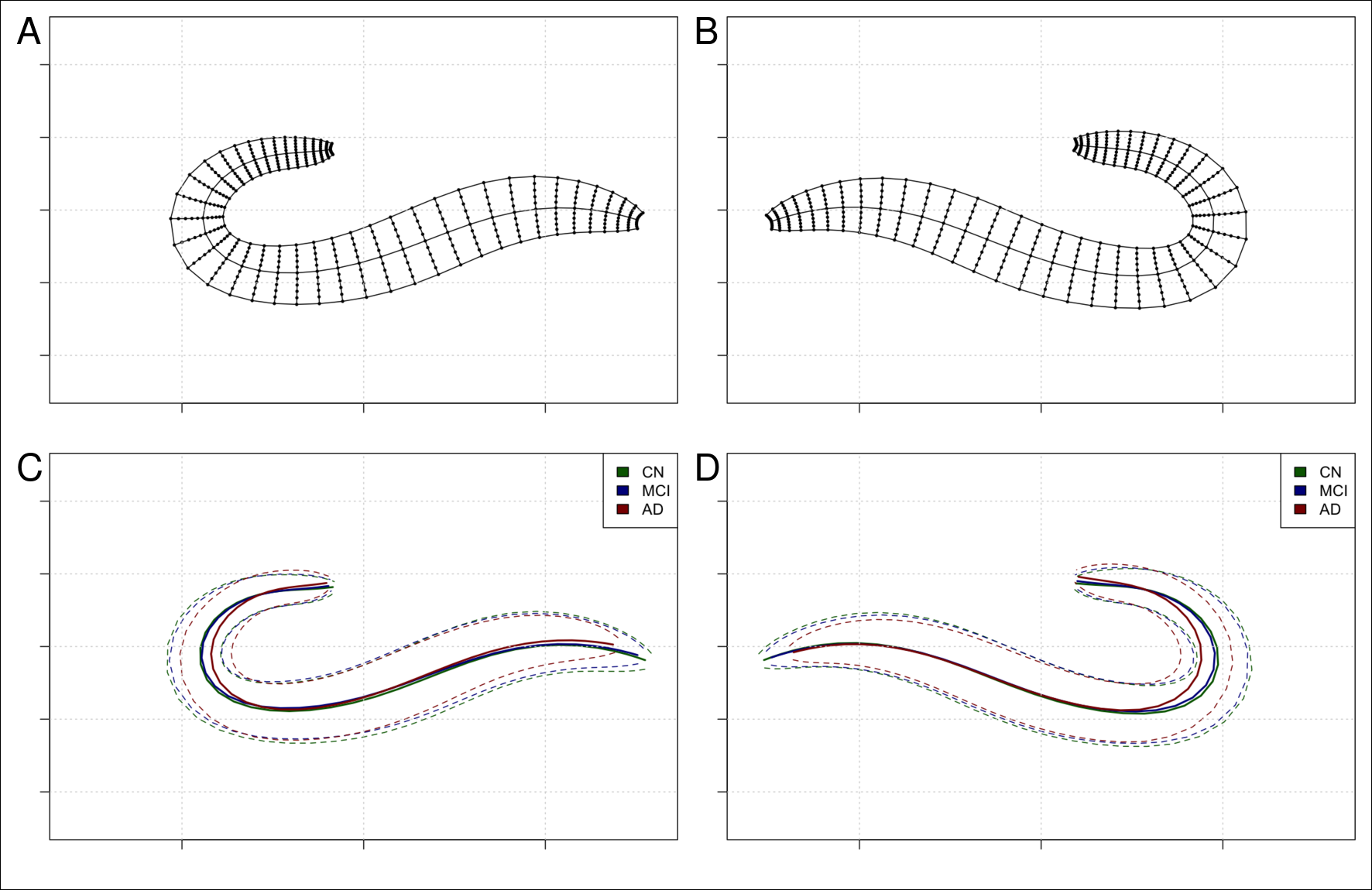}
    \end{framed}
    \captionsetup{width=0.475\textwidth}
    \caption{Slices through the hippocampus. Panels (a) and (b) show left and right thickness measurements at one particular location on the longitudinal axis. Panels (c) and (d) show shapes of the inner, outer, and mid-surface (group averages for cognitively unimpaired controls, mild cognitive impairment, and dementia groups from the DELCODE / ASHS dataset).}
    \label{fig:m-8}
\end{figure}

The hippocampal shape and thickness analysis provides a set of uni- and multivariate morphometric features, including both spatially localized as well as summary measures. The primary outcome are local thickness estimates, defined as the distance measured along the streamlines in the interior/exterior direction of the hippocampal shape model. For our experiments and analyses, we typically use a $40 \times 20$ coordinate grid in the medial/lateral and anterior/posterior directions, but in principle, arbitrary resolutions are possible. The values obtained at the grid points can then be visualized on either the hippocampal mid-surface, or on a rectangular plane, i.e.\ a 2D image such as in Figure~\ref{fig:m-7}.
Thickness, however, is just one instance of length-based measurements, since these measurements need not necessarily follow the interior/exterior direction, but can be done in the anterior/posterior and medial/lateral dimensions as well. This would give the spatial extent of the hippocampus in these dimensions. Further, distance measurements need not necessarily follow the streamlines at all, but can also be taken along the surface of the mesh, which provides circumference measures rather than thickness measures. For all of the above measurements, both uni- and multivariate versions exist, providing either a concise summary measure or detailed localization information, depending on the goal of the analysis.
While all length-based measurements are one-dimensional quantities, the shape model also allows to derive two- or three-dimensional quantities, such as the areas of a set of hippocampal slices at different locations (Figure~\ref{fig:m-8} (a) and (b)), or a model-based estimate of hippocampal volume, as opposed to a crude voxel-based measure. Also, two or more measures derived from the shape model can be combined into composite measures that may reveal additional morphometric changes, such as the ratio between inner and outer surface area, or the shape index, i.e.\ the ratio between circumference and surface area, or the surface area and volume. Lastly, the shape model also provides a means for mapping of data from other modalities - such as fMRI or PET signals - to the hippocampal mid-surface. 
In addition, while all of the above measures rely on length, area, or volume information, hippocampal geometry can also be characterized by its curvature, and changes in curvature may as well be indicative of shape changes across time or in health vs.\ disease (Figure~\ref{fig:m-8} (c) and (d)). Therefore, we also provide estimates of mean curvature at every coordinate of the hippocampal sheet (Figure~\ref{fig:m-5} (d)) and as summary measures along a given axis of the hippocampus (Figure~\ref{fig:m-6} (a) and (b)).

\subsection{Evaluation of the hippocampal shape and thickness analysis}

We evaluate the performance of our algorithm with a series of experiments and analyses in the domain of neurodegenerative changes in Mild Cognitive Impairment (MCI) and dementia of the Alzheimer type (DAT) as compared to cognitively unimpaired (CU) controls. For this purpose, we use two datasets in combination with two image preprocessing algorithms. 
Our primary dataset for the evaluation originates from the DZNE-Longitudinal Cognitive Impairment and Dementia Study (DELCODE) study \cite{jessen2018}. DELCODE is an ongoing, longitudinal multicentric imaging study in Alzheimer's disease, with an emphasis on its early, preclinical stages. At each DELCODE site, the local institutional review boards approved the study protocol, and the ethics committees issued local ethics approval. The study protocol followed the ethics principles for human experimentation in accordance with the Declaration of Helsinki. All participants in the study provided written informed consent. 
From this dataset, we used 3T, high-resolution (0.5 $\times$ 0.5 $\times$ 1.5 $mm^3$) T2-weighted images in conjunction with standard T1-weighted images. Details of the demographic and clinical characteristics of the analysis samples are given in Table~\ref{tab-dmgr}.

The second dataset used for evaluation was obtained from the Alzheimer’s Disease Neuroimaging Initiative (ADNI; \url{https://adni.loni.usc.edu}) database. ADNI is an ongoing multicentric, longitudinal imaging study on Alzheimer's disease and its prodromal syndrome, Mild Cognitive Impairment \cite{mueller2005, jack2008}. All participants provided written informed consent according to the Declaration of Helsinki and the study was approved by the institutional review board at each participating site. We included all cases for which high-resolution hippocampal T2-weighted images in combination with the corresponding T1-weighted images were available. Although acquisition parameters are heterogeneous in the ADNI study (e.g.\, due to scanner differences), all T2 images were acquired at 3T, had a resolution of (0.4 $\times$ 0.4 $\times$ 2.0 $mm^3$) or higher, and covered the whole hippocampus. Only a single time-point per participant was retained if longitudinal measurements were available. Again, details of the demographic and clinical characteristics of the analysis samples are given in Table~\ref{tab-dmgr}.

\begin{table}[ht]
\centering
\begingroup\footnotesize
\begin{tabularx}{0.475\textwidth}{L{1.25}L{0.35}R{1.75}R{1.15}R{0.5}}
\hline
& & \textbf{age [years] Mean($\pm$SD)} & \textbf{proportion female} & \textbf{n} \\ 
\hline
\multirow{4}{*}{DELCODE}
& CU & 69.4 ($\pm$5.36) & 0.58 & 221 \\ 
& MCI & 73.0 ($\pm$5.68) & 0.49 & 154 \\ 
& DAT & 74.8 ($\pm$6.46) & 0.60 &  93 \\ 
\cline{2-5}
& Total & 71.6 ($\pm$6.12) & 0.54 & 468 \\ 
\hline
\multirow{4}{*}{ADNI} 
& CU & 73.0 ($\pm$7.22) & 0.59 & 560 \\ 
& MCI & 73.8 ($\pm$7.88) & 0.43 & 426 \\ 
& DAT & 76.0 ($\pm$8.85) & 0.42 & 168 \\ 
\cline{2-5}
& Total & 73.7 ($\pm$7.78) & 0.51 & 1154 \\ 
\hline
\end{tabularx}
\endgroup
\captionsetup{width=0.475\textwidth,aboveskip=3pt,belowskip=6pt}
\caption{Demographic and clinical characteristics for the analysis samples from the DELCODE and ADNI studies. Abbreviations: CU=cognitively unimpaired controls, MCI=Mild Cognitive Impairment, DAT=dementia of the Alzheimer type.}
\label{tab-dmgr}
\end{table}

All images were processed with the ASHS software package \cite{yushkevich2015}, version 2.0.0, and the Penn ABC-3T ASHS Atlas for T2-weighted MRI \cite{xie2023deep}, to create segmentations of the hippocampus and its subfields. Both the high-resolution, hippocampal T2 image and the standard T1 image were processed. For our analyses, we retained the labels for the subiculum, CA1, CA2, and CA3 (note that the presubiculum is not included as a separate label in the ASHS segmentation, and that the subiculum in the ASHS segmentation overlaps with both the presubiculum and the subiculum in the FreeSurfer segmentation). All masks in the DELCODE/ASHS dataset were visually inspected for segmentation errors and manual edits were performed, where necessary and possible, to correct such errors (see Appendix \ref{sec:qc} for details). In addition, to evaluate the impact of image preprocessing strategy, we also used Freesurfer's hippocampal subfields segmentation \cite{iglesias2015} as implemented in Freesurfer 7.1.1, again utilizing the high-resolution T2 and the standard T1 image. Labels for the presubiculum, subiculum, CA1, CA2/3, and the molecular layer were used within this study. For both, the ASHS and Freesurfer results, we then apply our algorithm, including the curvature alignment, to compute thickness and further geometrical features of the hippocampus. Details about QC procedures and processing results are given in Appendix \ref{sec:qc}. For comparison, we also evaluate the results of the \textit{HippUnfold} algorithm (\url{https://github.com/khanlab/hippunfold}; \cite{dekraker2018, dekraker2020hippocampal}). We used the singularity container for \textit{HippUnfold} version 1.2, and processed the high-resolution hippocampal T2 image in conjunction with the standard T1 image for template registration.

Our primary goal in the following analyses is to illustrate our algorithm in a typical application scenario, to replicate well-known clinical group differences, and to evaluate whether or not the proposed tools provide additional sensitivity beyond the traditional, voxel-based measurements of hippocampal volume for distinguishing between clinical, subclinical, and cognitively unimpaired groups. While our algorithm also allows for other analyses such as the investigation of associations with cognitive measures or CSF or peripheral biomarkers, these are beyond the scope of this methods-oriented work.

\section{Results}

\subsection{Hippocampal volume}

We first examine group differences in ASHS-derived estimates of bilateral hippocampal volume as a benchmark for the subsequent thickness analyses. In a regression analysis with clinical group as the predictor of interest, and age, gender, and total intracranial volume as regressors of no interest, both the MCI and DAT groups show significant volume reductions in comparison to the control group (MCI $<$ controls: $t=8.23$, $p<0.001$; DAT $<$ controls: $t=13.44$, $p<0.001$). The group differences for total hippocampal volume are also reflected at the more detailed level of individual hippocampal subfields (Table~\ref{tab-hc-sub}). The overall pattern of larger volume losses in the DAT group than the MCI group remains the same, but group differences appear to be more pronounced towards the medial end of the hippocampus, i.e.\ rather in the subiculum or CA1 than in CA2 or CA3.

\begin{table}[ht]
\centering
\begin{subtable}{0.475\textwidth}
\centering
\begingroup\footnotesize
\begin{tabularx}{\textwidth}{L{1.6}L{0.8}R{0.8}R{0.8}}
\hline
& & {\textbf{$t$}} & {\textbf{$p$}} \\ 
\hline
\multirow{2}{*}{Subiculum} 
& DAT$<$CU & $ 9.64$ & $ <0.001$ \\ 
& MCI$<$CU & $ 5.16$ & $ <0.001$ \\ 
\hline
\multirow{2}{*}{CA1} 
& DAT$<$CU & $ 11.22$ & $ <0.001$ \\ 
& MCI$>$CU & $ 6.27$ & $ <0.001$ \\ 
\hline
\multirow{2}{*}{CA2} 
& DAT$<$CU & $ 6.18$ & $ <0.001$ \\ 
& MCI$>$CU & $ 2.21$ & $  0.027$ \\ 
\hline
\multirow{2}{*}{CA3} 
& DAT$<$CU & $ 2.69$ & $ 0.007$ \\ 
& MCI$>$CU & $ 0.78$ & $ 0.437$ \\ 
\hline
\end{tabularx}
\endgroup
\end{subtable}
\captionsetup{width=0.475\textwidth,aboveskip=3pt,belowskip=6pt}
\caption{Statistical evaluation of group differences for ASHS-derived estimates of selected hippocampal subfield volumes, controlling for age, sex, and total intracranial volume.}
\label{tab-hc-sub}
\end{table}

\subsection{Hippocampal thickness}

\begin{figure*}[!hbt]
\centering
\begin{subfigure}[b]{0.95\textwidth}
    \includegraphics[width=\textwidth]{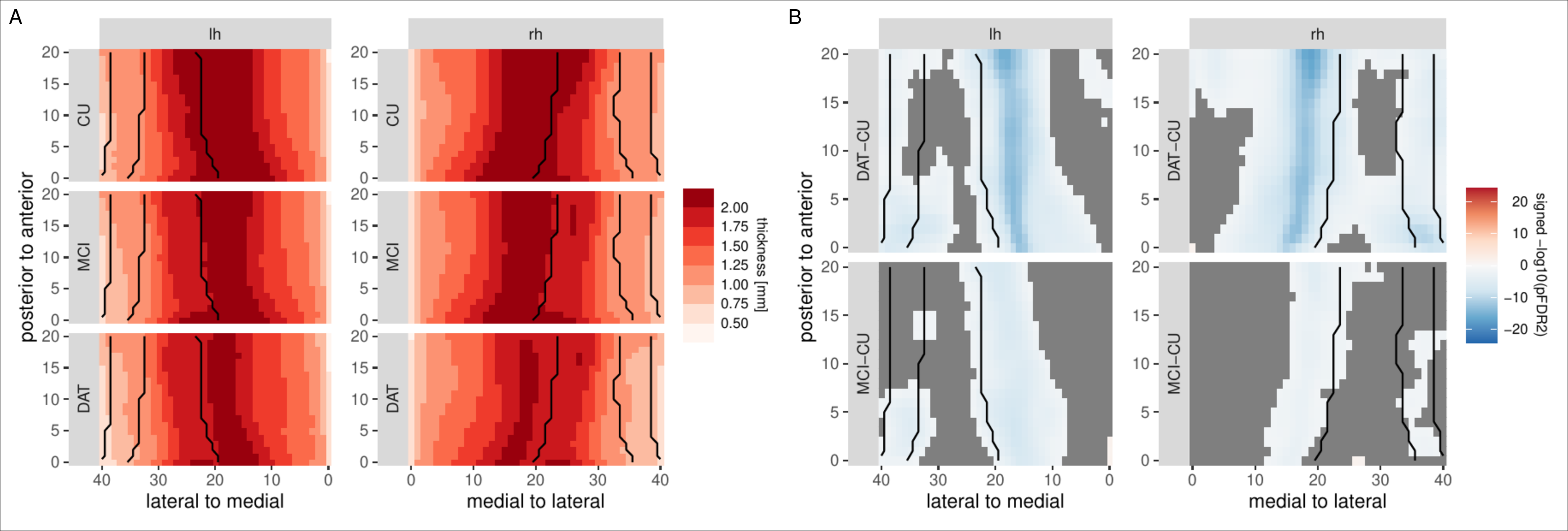}
\end{subfigure}
\captionsetup{width=0.95\textwidth}
\caption{Evaluation of group differences in hippocampal thickness, based on an ASHS segmentation in the DELCODE sample. Left: mean localized hippocampal thickness estimates for the left and right hemisphere in the DAT, MCI, and control groups. Right: statistical evaluation of thickness differences between the controls and the MCI and DAT groups. Orange/blue colors indicate regions that are significant after FDR2-correction for multiple comparisons. Black lines indicate, from medial to lateral, boundaries between the subiculum, CA1, CA2, and CA3.}
\label{fig:r-3}
\end{figure*}

We next evaluate our algorithm with respect to the spatial distribution of localized thickness estimates and its ability to reveal differences between clinical groups. Figure~\ref{fig:r-3} shows raw hippocampal thickness in the DAT, MCI, and control groups and the result of a regression model with the predictors group, age, and gender that was fit at each vertex. Hippocampal thickness primarily varies along the medial-lateral axis, and highest values are observed in CA1 and the subiculum. The differences in thickness estimates between the diagnostic groups reflect and add more detail to the general pattern of group differences observed for total hippocampal volume and the volumes of hippocampal subfields: higher reductions in thickness as compared to the control group are observed in the DAT group than in the MCI group, and both remain significant after correction for multiple comparisons (Figure~\ref{fig:r-3} and Table~\ref{tab-r-ht-peaksInt}). Although wide-spread, these differences are not uniform across the hippocampal body, but vary along its extent, primarily across the medial-lateral axis, and less across the anterior-posterior axis, with most pronounced differences being observed at the border between CA1 and the subiculum. 

\begin{table}[ht]
\centering
\begin{subtable}{0.475\textwidth}
\centering
\begingroup\footnotesize
\begin{tabularx}{\textwidth}{L{1.6}C{0.6}C{0.6}C{0.6}R{0.8}R{1.4}R{1.4}}
\hline
& & {\textbf{x}}  & {\textbf{y}} & {\textbf{$t$}} & {\textbf{$p$}} & {\textbf{$p_{FDR2}$}} \\ 
\hline
DAT$<$CU  & left  & 19 & 21 & $9.09$ & $<0.001$ & $<0.001$ \\ 
DAT$<$CU  & right & 20 & 21 & $9.53$ & $<0.001$ & $<0.001$ \\ 
MCI$<$CU & left  & 18 &  5 & $6.60$ & $<0.001$ & $<0.001$ \\ 
MCI$<$CU & right & 20 & 21 & $5.44$ & $<0.001$ & $<0.001$ \\ 
\hline
\end{tabularx}
\endgroup
\end{subtable}
\captionsetup{width=0.475\textwidth,aboveskip=3pt,belowskip=6pt}
\caption{Peak coordinates and statistics for the evaluation of aligned thickness estimates. For all analyses, clinical group was used as predictor of interest, and age and sex as covariates of no interest.}
\label{tab-r-ht-peaksInt}
\end{table}

\subsection{Hippocampal geometry}\label{sec:geometry}

%\subsubsection{Length-based measures}

A straightforward extension of the proposed thickness measurements is to measure distance not only in the $z$ direction, but also in the $x$ and $y$ coordinate directions. For clarity, we will refer to these two measurements as measures of \emph{extent}, whereas we reserve the term \emph{thickness} to measurements in the $z$ direction exclusively. Table~\ref{tab-r-ht-lengths} shows the statistical evaluation of group differences for geometry-based summary measures of hippocampal extent and thickness, averaged across hemispheres. For the MCI and DAT groups, reductions in extent compared to controls are observed in the medial-lateral ($x$), but not the anterior-posterior ($y$) direction. For both groups, reductions are also present for the mean hippocampal thickness ($z$) summary measure, as could be expected based on the previously observed reductions in the localized thickness estimates (cf.\ Figure~\ref{fig:r-3}). 

\begin{table}[ht]
\centering
\begin{subtable}{0.475\textwidth}
\centering
\begingroup\footnotesize
\begin{tabularx}{\textwidth}{L{1.6}L{0.8}R{0.8}R{0.8}}
\hline
& & {\textbf{$t$}} & {\textbf{$p$}} \\ 
\hline
\multirow{2}{*}{Length (x)} 
& DAT$<$CU & $8.91$ & $<0.001$ \\ 
& MCI$<$CU & $4.74$ & $<0.001$ \\ 
\hline
\multirow{2}{*}{Length (y)} 
& DAT$<$CU & $1.01$ & $0.311$ \\ 
& MCI$<$CU & $-0.08$ & $0.939$ \\ 
\hline
\multirow{2}{*}{Length (z)} 
& DAT$<$CU  & $8.95$ & $<0.001$ \\ 
& MCI$<$CU & $5.13$ & $<0.001$ \\ 
\hline
\end{tabularx}
\endgroup
\end{subtable}
\captionsetup{width=0.475\textwidth}
\caption{Statistical evaluation of group differences for summary measures of hippocampal extent and thickness. For all analyses, clinical group was used as predictor of interest, and age and sex as covariates of no interest.}
\label{tab-r-ht-lengths}
\end{table}

%\subsubsection{Curvature-based measures}

\begin{figure*}[!hbt]
\centering
\begin{subfigure}[b]{0.95\textwidth}
    \includegraphics[width=\textwidth]{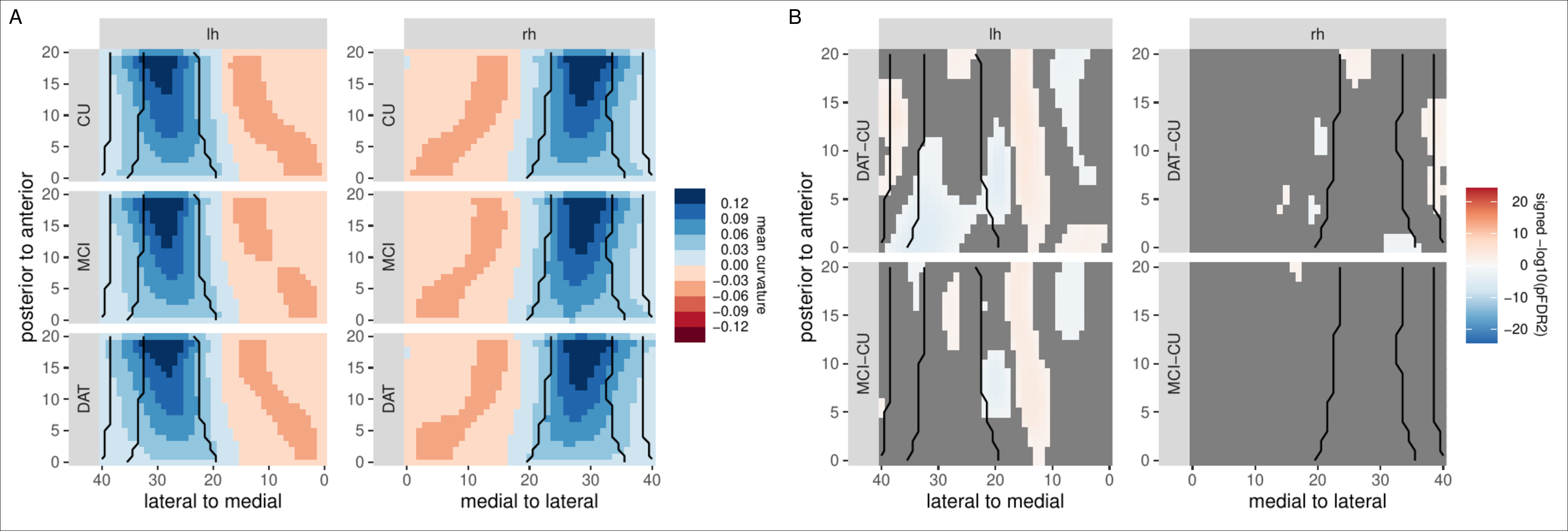}
\end{subfigure}
\captionsetup{width=0.95\textwidth}
\caption{Evaluation of group differences in curvature, based on an ASHS segmentation in the DELCODE sample. Left: localized curvature estimates for the left and right hemisphere in the DAT, MCI, and control groups. Right: statistical evaluation of curvature differences between the controls and the MCI and DAT groups. Orange/blue colors indicate regions that are significant after FDR2-correction for multiple comparisons. Black lines indicate, from medial to lateral, boundaries between the subiculum, CA1, CA2, and CA3.}
\label{fig:r-4}
\end{figure*}

Curvature is a geometric measure complementary to length-based measurements. Here, we evaluate its potential to reveal additional shape characteristics beyond hippocampal thickness. Figure~\ref{fig:r-4} shows localized mean curvature estimates in the left and right hemisphere for the three diagnostic groups. The overall pattern is similar across groups, with a pronounced bend towards the lateral end of the hippocampus, and a less pronounced one towards its medial end. A statistical comparison shows curvature differences between the DAT and CU as well as the MCI and CU groups, primarily in the left hemisphere, with a more pronounced curvature increase in medial regions.

%\subsubsection{Circumference- and area-based measures}

Even beyond localized hippocampal thickness and curvature, the geometry-based representation of the hippocampus allows to derive additional measures such as a) the circumference, defined as the sum of inner/outer lengths, b) the ratio of these lengths, called the interior/exterior ratio, c) the surface area, and d) the 2D shape index, which is the ratio of circumference and surface area. Here, we explore if these measures can explain differences in hippocampal shape in addition to volume, thickness, and curvature. We observe group differences between the DAT and MCI groups and the control group, respectively, for hippocampal circumference and surface area, but not for interior/exterior ratio, nor for the shape index (Table~\ref{tab-r-geom}).

\begin{table}[ht]
\centering
\begin{subtable}{0.475\textwidth}
\centering
\begingroup\footnotesize
\begin{tabularx}{\textwidth}{L{1.6}L{0.8}R{0.8}R{0.8}}
\hline
& & {\textbf{$t$}} & {\textbf{$p$}} \\ 
\hline
\multirow{2}{*}{Circumference} 
& DAT$<$CU  & $9.18$ & $<0.001$ \\ 
& MCI$<$CU & $4.88$ & $<0.001$  \\ 
\hline
\multirow{2}{*}{Interior/exterior ratio} 
& DAT$<$CU  & $-0.18$ & $0.856$ \\ 
& MCI$<$CU & $0.57$ & $0.571$  \\ 
\hline
\multirow{2}{*}{Surface area} 
& DAT$<$CU  & $7.46$ & $<0.001$  \\ 
& MCI$<$CU & $3.27$ & $ 0.001$  \\ 
\hline
\multirow{2}{*}{Shape index} 
& DAT$<$CU  & $-1.40$ & $0.163$ \\ 
& MCI$<$CU & $0.06$ & $0.951$  \\ 
\hline
\end{tabularx}
\endgroup
\end{subtable}
\captionsetup{width=0.475\textwidth,aboveskip=3pt,belowskip=6pt}
\caption{Statistical evaluation of group differences for hippocampal circumference estimates, interior/exterior ratio, surface area, shape index (circumference/area). For all analyses, clinical group was used as predictor of interest, and age and sex as covariates of no interest.}
\label{tab-r-geom}
\end{table}

\subsection{Classification}

We next evaluate whether hippocampal thickness has an incremental explanatory value for the classification of the MCI and DAT groups vs.\ cognitively unimpaired controls. Specifically, we evaluate if the addition of hippocampal thickness information improves the performance (area under the curve, AUC) of a logistic regression model that contains hippocampal volume as the only other predictor. Statistical significance of improvements is assessed using the likelihood ratio test. Table~\ref{tab-r-class-zInt} shows that classification performance is higher for the DAT group than for the MCI group. Furthermore, the addition of thickness information improves the performance for both groups -- even when using only mean hippocampal thickness as a broad summary measure -- indicating the complementary information in these measurements. In contrast to hippocampal thickness, the geometry-based measures do not provide incremental explanatory value beyond hippocampal volume (see Section~\ref{sec:incr} in the Appendix).

\begin{table}[ht]
\centering
\begin{subtable}{0.475\textwidth}
\centering
\begingroup\footnotesize
\begin{tabularx}{\textwidth}{L{1.1}R{0.9}C{1.3}R{0.7}}
\hline
& {\textbf{volume}} & {\textbf{volume \& thickness}} & {\textbf{LR test}} \\ 
\hline
DAT vs.\ CU & $0.88\pm 0.05$ & $0.90\pm 0.05$ & $<0.001$ \\ 
MCI vs.\ CU & $0.73\pm 0.04$ & $0.75\pm 0.06$ & $0.007$ \\ 
\hline
\end{tabularx}
\endgroup
\end{subtable}
\captionsetup{width=0.475\textwidth,aboveskip=3pt,belowskip=6pt}
\caption{Classification performance (AUC$\pm$CI) and $p$-values of the likelihood ratio (LR) test for hippocampal volume and hippocampal thickness.}
\label{tab-r-class-zInt}
\end{table}

\subsection{Replication with different datasets and segmentation algorithms}\label{sec:replication}

\begin{figure*}[!hbt]
\centering
\begin{subfigure}[b]{0.95\textwidth}
    \includegraphics[width=\textwidth]{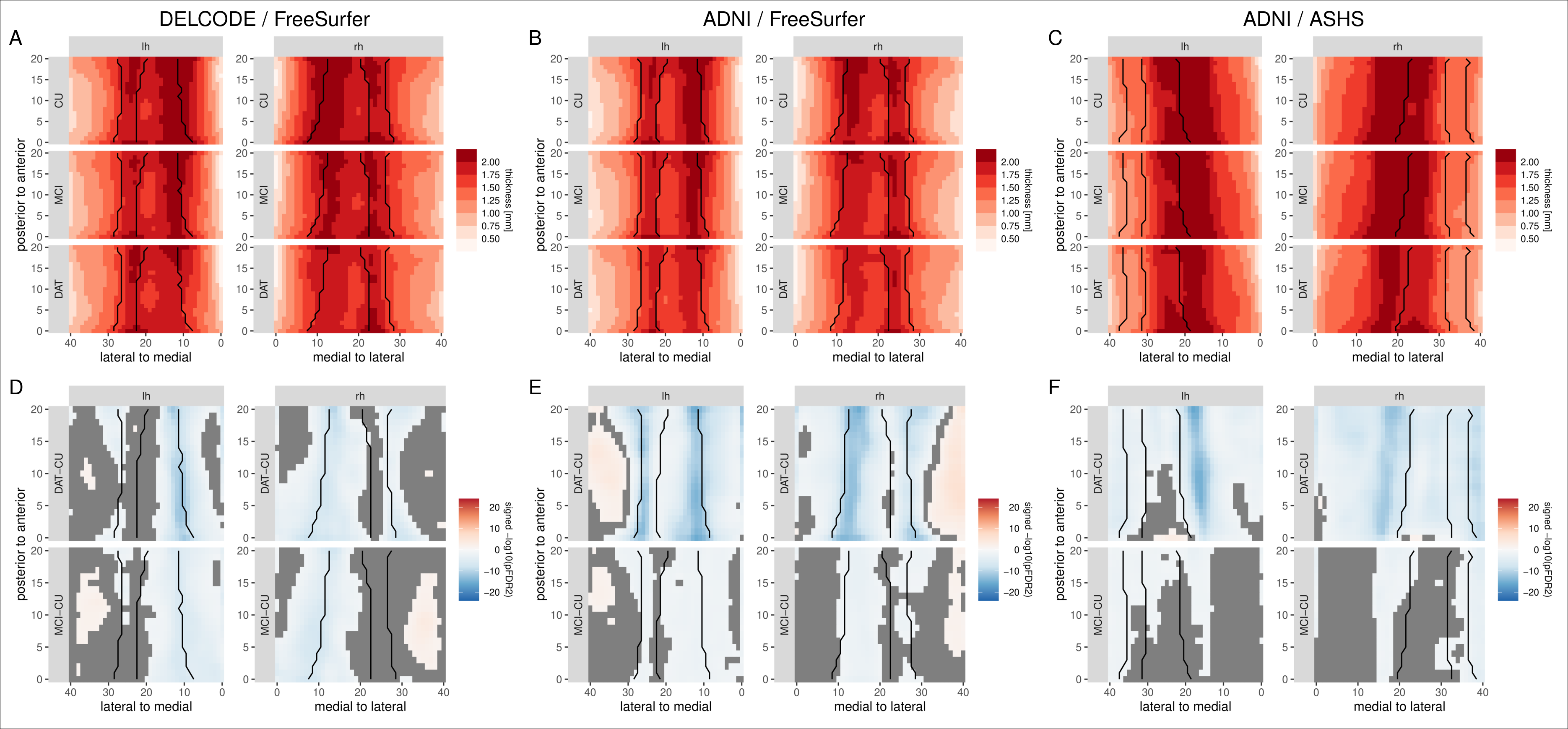}
\end{subfigure}
\captionsetup{width=0.95\textwidth}
\caption{Localized thickness per diagnostic group and statistical evaluation of group differences for different datasets and segmentation algorithms. Left: DELCODE / FreeSurfer, middle: ADNI / FreeSurfer, right: ADNI / ASHS. Note, that the ASHS and FreeSurfer results cannot be compared directly due to differences in the underlying segmentation protocols.}
\label{fig:r-6}
\end{figure*}

Finally, we evaluate the ability of the algorithm to handle different datasets and different image segmentation algorithms. Specifically, we attempt to replicate the main results using FreeSurfer's hippocampal subfields segmentation algorithm (Figure~\ref{fig:r-6}(a) and \ref{fig:r-6}(d)). Further, we also exchange the dataset and re-run the analysis for the ADNI subset with both the FreeSurfer (Figure~\ref{fig:r-6}(b) and \ref{fig:r-6}(e)) and the ASHS (Figure~\ref{fig:r-6}(c) and \ref{fig:r-6}(f)) segmentation algorithm. The overall pattern of results is similar across algorithms and datasets, and consistent with the main results obtained from the DELCODE data and ASHS segmentation: the highest thickness values are observed in the subiculum, with little variation along the anterior-posterior axis. For all analyses, these estimates show decreases for the MCI and DAT groups as compared to the control group, with more pronounced decreases in the DAT than in the MCI group. The statistical evaluation shows that these differences are significant primarily in the subiculum and CA1. The most lateral regions of the hippocampus show, in contrast, an increase in thickness, at least for the FreeSurfer analyses. A supplemental analysis with a purely image-based thickness estimation algorithm (Section~\ref{sec:increase} in the Appendix) indicates that these are likely already present in the segmentation images and not introduced as an artifact of the proposed thickness estimation method. In contrast to the main results in the DELCODE / ASHS data, an analysis of the incremental validity of the hippocampal thickness estimates shows an added value only for MCI vs.\ CU classification in the DELCODE / FreeSurfer data and for the DAT vs.\ CU classification in the ADNI / ASHS data (Table~\ref{tab-r-class-zInt-others}).

\begin{figure*}[!hbt]
\centering
\begin{subfigure}[b]{0.95\textwidth}
    \includegraphics[width=\textwidth]{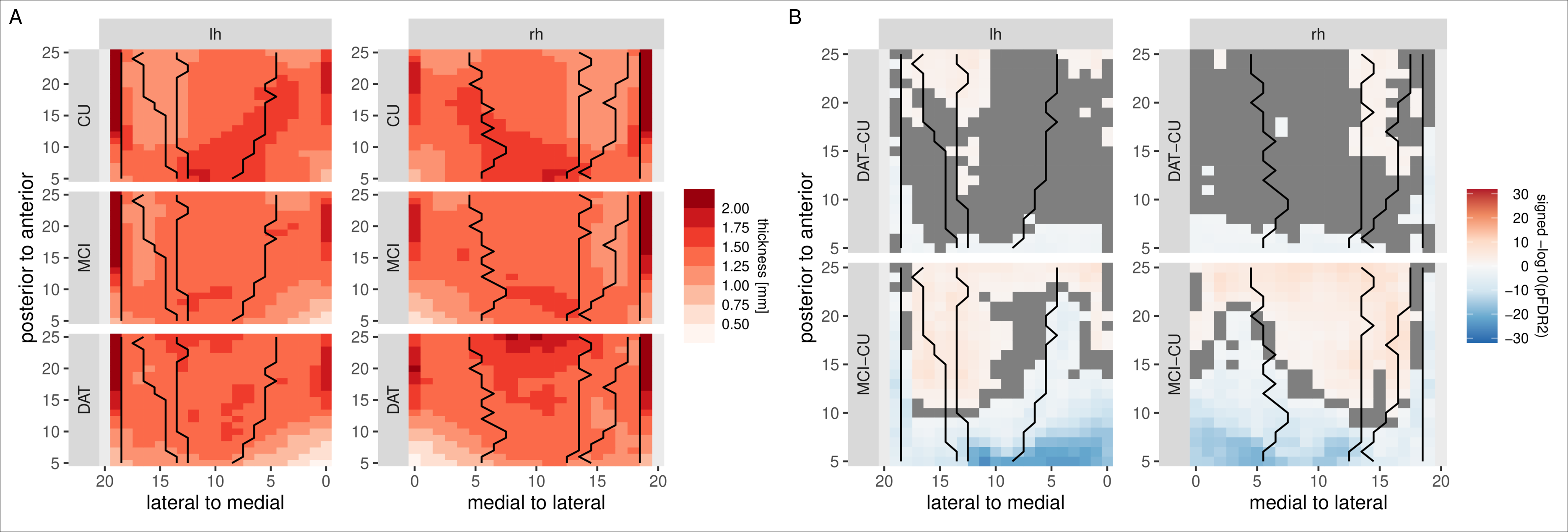}
\end{subfigure}
\captionsetup{width=0.95\textwidth}
\caption{Evaluation of group differences in hippocampal thickness, based on the \textit{HippUnfold} algorithm in DELCODE. The figures have been cropped approximately to the hippocampal body region. Left: mean localized hippocampal thickness estimates for the left and right hemisphere in the DAT, MCI, and cognitively unimpaired control groups. Right: statistical evaluation of thickness differences between the controls and the MCI and DAT groups. Orange/blue colors indicate regions that are significant after FDR2-correction for multiple comparisons. Black lines indicate subfield boundaries between CA4, CA3, CA2, CA1, and subiculum, from lateral to medial. Note, that there is no correspondence between coordinates for the \textit{HippUnfold} algorithm and the algorithm proposed in this paper.}
\label{fig:r-7}
\end{figure*}

Figure~\ref{fig:r-7} depicts the results of the \textit{HippUnfold} algorithm on the DELCODE dataset. For better comparison with the above results in the hippocampal body, we here restrict the comparison to this region. However, since the \textit{HippUnfold} algorithm not only unfolds the body, but the complete hippocampus, we also show the full extent of the unfolding in Appendix \ref{sec:hu-delcode-full}. We observe that thickness is highest in the most lateral regions in all groups. The statistical evaluation highlights that thickness decreases are primarily present in posterior regions of the hippocampus in the DAT group, and to a lesser extent in the MCI group. Further, thickness increases appear to be present in CA1, CA2 and the subiculum. Similar results are obtained for an analysis of the ADNI dataset (see Section~\ref{sec:hu-adni-full} in the Appendix).

\begin{table*}[ht]
\centering
\begin{subtable}{0.75\textwidth}
\centering
\begingroup\footnotesize
\begin{tabularx}{\textwidth}{L{1.4}L{1}R{0.7}R{1.2}R{0.7}}
\hline
& & {\textbf{volume}} & {\textbf{volume+thickness}} & {\textbf{LR test}} \\ 
\hline
\multirow{2}{*}{DELCODE / FreeSurfer}
& DAT vs.\ CU & $0.88\pm 0.07$ & $0.89\pm 0.06$ & $0.184$ \\ 
& MCI vs.\ CU & $0.75\pm 0.07$ & $0.76\pm 0.07$ & $0.024$ \\
\hline
\multirow{2}{*}{ADNI / FreeSurfer} 
& DAT vs.\ CU & $0.87\pm 0.04$ & $0.87\pm 0.03$ & $0.104$ \\ 
& MCI vs.\ CU & $0.61\pm 0.04$ & $0.61\pm 0.07$ & $0.131$ \\ 
\hline
\multirow{2}{*}{ADNI / ASHS}
& DAT vs.\ CU & $0.84\pm0.06$ & $0.86\pm0.05$ & $0.005$ \\
& MCI vs.\ CU & $0.62\pm0.05$ & $0.62\pm0.05$ & $0.197$ \\
\hline
\end{tabularx}
\endgroup
\end{subtable}
\captionsetup{width=0.75\linewidth}
\caption{Classification performance (AUC$\pm$CI) and $p$-values of the likelihood ratio (LR) test for hippocampal volume and hippocampal thickness using different datasets and segmentation algorithms.}
\label{tab-r-class-zInt-others}
\end{table*}

%\FloatBarrier

%\newpage
\section{Discussion}

% summary of research
In this work, we have presented an approach for the unfolding of the hippocampus, for the creation of a common space and coordinate system across individuals, for thickness estimation, and the extraction of shape features. We have further conducted an extensive empirical analysis to evaluate the proposed algorithm in a range of prototypical application scenarios.

% summary & interpretation of main results
Our main results can be summarized as follows: first, in comparison with traditional volumetric estimates, localized thickness estimates provide additional information in their ability to pinpoint the location of effects onto the hippocampal sheet. In this regard, we observed that thickness differences were most pronounced at the border between CA1 and the subiculum. While this is in line with our observations of greater volumetric differences in the subiculum and CA1 than in CA2 or CA3, our method adds spatial detail to this observation. We take this as an indication of the utility of our algorithm.
Hippocampal thickness estimates also provide additional, complementary information in their ability to augment existing analyses. This is evident from the improvement in classification performance when thickness was added as an additional predictor. In that sense, hippocampal thickness captures information that is not redundant with or included in traditional estimates. Therefore, the incremental value beyond traditional analysis strategies is another indication of the utility. 
Second, hippocampal thickness differentiates between clinical groups consistent with traditional, volume-based estimates. Specifically, we have been able to replicate group differences between the DAT and MCI groups and the control group, respectively, for hippocampal thickness, with larger differences for the DAT vs.\ CU comparison than for the MCI vs.\ CU comparison. The reproducibility of these known effects points to the validity of our algorithm.
Third, our proposed method for the analysis of hippocampal thickness works across different datasets and can be used with different segmentation algorithms. Specifically, we obtain similar, although not identical, results for the ADNI and DELCODE data as well as for the FreeSurfer and ASHS segmentation tools. With regard to group differences and spatial distributions of effects, the main results hold for all combinations of datasets and segmentation algorithms, emphasizing the generalizability of our algorithm.

% discussion of notable results
We have observed a set of notable results that deserve further discussion: first, the vertex-wise results indicate that there are indeed effects that extend beyond the boundaries of a single subfield and/or do not cover a single subfield entirely. This makes the detection of these effects with traditional region-based approaches difficult. It also underlines the importance of localized methods that are independent of potentially unreliable boundaries between hippocampal subfields and instead permit the detection and localization of effects extending across subfield boundaries. Further, localized approaches, such as ours, permit even targeted analyses with custom region of interest definitions that can focus on a dedicated hypothesis. 

Second, the analysis of geometric features beyond thickness has demonstrated that traditional shape metrics such as shape index or interior/exterior ratio, which are based on global shape characteristics, are limited in revealing group differences. This underlines the need for novel and detailed shape descriptors such as localized hippocampal thickness. At the same time, significant group differences in measures such as lateral/medial extent, curvature, circumference, or surface area illustrate that anatomical changes of the hippocampus in MCI and Alzheimer's disease dementia are multi-dimensional, and cannot be solely captured by a single summary metric such as volume. Whether or not these measures can improve existing analyses remains an open question, given that we did not observe incremental explanatory value beyond hippocampal volume in our analyses. However, this does not rule out that they could be relevant in other application contexts (e.g., other diseases).

Third, differences and commonalities between datasets and segmentation algorithms have likely had an impact on the results of our algorithm. While a comparison was not the primary focus of this study, we note that ADNI is a larger and more heterogeneous study than DELCODE, with a higher number of sites and a greater heterogeneity with regard to, for example, imaging devices and acquisition regimes. The two studies also have different demographics and clinical characteristics. This suggests that a direct comparison can only be made with caution. In spite of these differences, the results for the ADNI vs.\ DELCODE samples appear to be relatively similar, and remaining differences appear to be driven rather by the underlying segmentation algorithm and corresponding region definition. Differences in overall thickness, for example, can be expected due to the inclusion of the molecular layer in the FreeSurfer segmentations, but not in the ASHS or HippUnfold segmentations. Nevertheless, the overall picture is that comparable results can be expected for the thickness estimation, but need to be interpreted with reference to the underlying segmentation algorithm. 

This is also true for the comparison with the \textit{HippUnfold} algorithm, which covers the complete Hippocampus and employs a different thickness estimation approach, resulting in only limited correspondence between the two algorithms. This is primarily due to the different underlying segmentations as well as different boundary definitions. With regard to the latter, the main difference is in the anterior/posterior dimension, where the HippUnfold algorithm includes the head and tail, while ours does not. Also in the laminar dimension, consistent thickness differences can be expected due to the inclusion of the molecular layer / SRLM in the FreeSurfer segmentation. In contrast, we assume more similarity with regard to the medial/lateral axis (i.e., proximal/distal boundaries), since the segmentations in both algorithms extend from the subiculum to CA3 (sometimes a small part of CA4 is also included in the HippUnfold segmentation). Despite these differences between the algorithms, there are some commonalities: Both showed more pronounced differences for the AD group than the MCI group when compared with the CU group, and both showed thickness decreases as well as increases in various parts of the hippocampus. However, the algorithms differ with regard to the localization of these effects: while our algorithm primarily detects thickness decreases in the MCI and AD groups in the subiculum/CA1 area along the anterior-posterior axis, the HippUnfold algorithm localizes these effects primarily along the whole medial-lateral axis and in the posterior part of the hippocampus. We also note a lesser amount of apparent increases in hippocampal thickness in our algorithm, and that differences between clinical groups appear to align more with subfield boundaries. It is difficult to speculate about the causes the observed differences. The HippUnfold algorithm and our algorithm are similar, and both share the same principal idea of applying differential geometry operators to MR images. Leaving the obvious difference of the inclusion of head and tail in the HippUnfold algorithm aside, there are differences in the underlying segmentations, in the post-processing of the segmentations, the creation of meshes, and also the unfolding procedure itself. For these reasons, differences between the two methods are to be expected.

Finally, the results of the classification analyses turned out to be mixed, especially when considering datasets and segmentations other than the DELCODE / ASHS combination. In particular for the ADNI / FreeSurfer, hippocampal thickness did not have an incremental predictive value beyond hippocampal volume for the classification between either the DAT or MCI groups and the control group. One potential explanation is the use of averaged hippocampal thickness as a summary measure, which neglects the regional specificity of thickness differences. Another explanation, at least for the DAT vs.\ controls classification, is the potential presence of ceiling effects; that is, an excellent discrimination between these groups can already be achieved with hippocampal volume alone, without much room for improvement.

% discussion of paradoxical increase
A particularly noteworthy observation was that surprising thickness increases appeared to be present in some regions of the hippocampus for some dataset and segmentation algorithm combinations. We conducted an additional experiment with a purely image-based thickness estimation algorithm (as opposed to our geometry-based algorithm) to determine if these unexpected effects could be due to our method, the hippocampal segmentation, or the image data itself (Appendix \ref{sec:increase}). We take the results as evidence that the increases are either due to the segmentation or even reflect actual anatomical differences. The ultimate cause of the observed effects cannot be determined in the present study. However, we speculate that volume and shape changes in Alzheimer's disease are more complex than simple uniform shrinkage. For example, the displacement or shape change of neighboring structures will lead to shape changes in the hippocampus, causing a deformation of this structure. Even if the general pattern, as indicated by both volume and thickness estimates, is still characterized by an overall decrease in volume and thickness, this does not rule out that there are locally heterogeneous effects, which can even take the form of local increase. We argue that if viewed as a deformation process, hippocampal shape analysis will particularly benefit from advanced methods that have sufficient sensitivity to reveal such subtle changes.

% discussion of curvature-based alignment
The curvature-based alignment of hippocampal thickness is intended to compensate for differences in the lateral-medial extent of the hippocampus that may bias the correspondence of coordinates across clinical groups. Since in its current form it is based on an average of curvature along the anterior-posterior dimension, this may not be fully accounting for overall shape differences. The proposed algorithm can in principle be used for alignment at each longitudinal coordinate. However, this has to be weighted against a potential increase of noise in the non-averaged curvature estimates.

% discussion of data quality issues and processing errors
We note, that the quality of the results depends on the data characteristics and the segmentation algorithm, i.e.\ some datasets and segmentations may be better suited than others for an analysis using our algorithm. Here, we specifically observed challenges with the ADNI data in conjunction with the ASHS segmentations. We speculate that this is for three reasons: first, ADNI data are characterized by highly anisotropic voxels ($2 mm$ maximum edge length). This makes the segmentations more prone for discontinuities, which increases the likelihood of holes or bridges. Second, ASHS does not internally upsample like the FreeSurfer segmentation method. This leads to anisotropic segmentations for ASHS, whereas the FreeSurfer segmentation outputs have $0.33 mm$ isotropic voxels, which eventually allow for smoother segmentations. Finally, for the ADNI/ASHS analysis, and in contrast to the other analyses, no explicit labels for the hippocampal head were available, which means that the boundary between hippocampal head and body needs to be estimated heuristically.

% limitations
A few limitations of our approach are also worth mentioning. In its current form, the algorithm is limited to an analysis of the hippocampal body, and does not include regions of the hippocampal head and tail. This is because automated segmentations and contemporary voxel-resolutions currently either do not provide sufficient detail such as in the tail of the hippocampus, or do not provide a segmentation that is easily suited for unfolding, such as in the hippocampal head (due to folding of the hippocampal head onto itself). However, if this changes with e.g.\ improved segmentation algorithms or improved resolution and contrast in MR images, our algorithm could be extended to include these regions as well. A second limitation is that our algorithm depends on the quality of the hippocampal sub-segmentations provided by e.g.\ FreeSurfer or ASHS. We observed that around 5\% of those segmentations contain topological defects, such as holes, and cannot be processed with our algorithm (it will terminate and report an error). Manual correction of the segmentations can -- in most cases -- fix those errors, if needed. It should also be acknowledged that some variance may also be introduced by the cutting planes that separate the hippocampal body from the head and tail. Also the placement of the mid-surfaces within the hippocampus can vary, depending on the construction of the 3D coordinate system. However, this does not necessarily impact the computation of the thickness estimates much, since these follow the streamlines through the coordinate system in the direction orthogonal to the mid-surface. Finally, the automated processing can fail, especially if the basic structure of the hippocampal body is not intact in the segmentation inputs. We, therefore, provide QC images and detailed error messages in case of failures, to support the user. We, however, expect that for reasonable segmentation quality, no manual intervention is needed for proper function of the algorithm.\color{black}

% outlook
The proposed algorithm offers several further analysis options that have not been evaluated in the present work. One example is the mapping of other signals such as PET or fMRI data onto the hippocampal sheet, which allows a precise localization and group comparison, as well as a straightforward correlation of these signals with hippocampal thickness data. Beyond that, an extension of the algorithm to regions beyond the hippocampus is possible: whenever boundaries can be defined and a 2D grid gives a reasonable characterization of the particular brain structure, our methodology is principally applicable to neighboring structures such as the entorhinal cortex as well, either in conjunction with the hippocampus, or as a separate entity. In addition, regions in the hippocampal head and tail could be included in the unfolding algorithm whenever advanced MR acquisition protocols -- ideally with isotropic voxel resolution -- give enough detail and a clear separation between the folding in these regions. Due to the similar folding structure in these regions, we speculate that the curvature-aware anisotropic Laplace-Beltrami operator will be useful for identifying anatomical landmarks in these regions as well. Another future extension concerns the incorporation of an equivolumetric approach for thickness estimation which can be fitted directly into the thickness direction: In fact, \cite{leprince2015combined} start with a Laplacian-based level-set definition (i.e.\ the function that we also estimate) and extend it via an advection approach. An advantage of this approach would be a compensation for curvature in creating a 3D coordinate system; it would, however, primarily affect the spacing and distances of layers within that coordinate system, but not change the thickness estimates that are the main variable of interest in the current work.

% conclusions
In this work, we have presented a novel algorithm to create a sheet representation and an intrinsic coordinate system of the hippocampal body. Our approach permits an unfolding of the hippocampus and the creation of a reference frame that is consistent across individual cases. This gives a point-wise correspondence of the hippocampal sheet across hemispheres and individuals. In addition to measures of hippocampal thickness, our approach allows for the analysis of geometric features that give additional information about hippocampal shape changes. Finally, in a series of evaluations, our algorithm has demonstrated its clinical utility, validity, and generalizability beyond traditional, voxel-based measurements of hippocampal volume.

\clearpage

\section*{Acknowledgements}

This work was supported by DZNE institutional funds, by the Federal Ministry of Education and Research of Germany (031L0206, 01GQ1801), and by NIH (R01 LM012719, R01 AG064027, R56 MH121426, and P41 EB030006).

Data used in preparation of this article were obtained from the DELCODE study. The DELCODE study was funded by the German Center for Neurodegenerative Diseases (Deutsches Zentrum für Neurodegenerative Erkrankungen, DZNE), reference number BN012. 

Data used in preparation of this article were also obtained from the Alzheimer’s Disease Neuroimaging Initiative (ADNI) database. As such, the investigators within the ADNI contributed to the design and implementation of ADNI and/or provided data but did not participate in analysis or writing of this report. A complete listing of ADNI investigators can be found at \url{https://adni.loni.usc.edu}. Data collection and sharing for this project was funded by the Alzheimer's Disease Neuroimaging Initiative (ADNI) (National Institutes of Health Grant U01 AG024904) and DOD ADNI (Department of Defense award number W81XWH-12-2-0012). ADNI is funded by the National Institute on Aging, the National Institute of Biomedical Imaging and Bioengineering, and through generous contributions from the following: AbbVie, Alzheimer’s Association; Alzheimer’s Drug Discovery Foundation; Araclon Biotech; BioClinica, Inc.; Biogen; Bristol-Myers Squibb Company; CereSpir, Inc.; Cogstate; Eisai Inc.; Elan Pharmaceuticals, Inc.; Eli Lilly and Company; EuroImmun; F. Hoffmann-La Roche Ltd and its affiliated company Genentech, Inc.; Fujirebio; GE Healthcare; IXICO Ltd.; Janssen Alzheimer Immunotherapy Research \& Development, LLC.; Johnson \& Johnson Pharmaceutical Research \& Development LLC.; Lumosity; Lundbeck; Merck \& Co., Inc.; Meso Scale Diagnostics, LLC.; NeuroRx Research; Neurotrack Technologies; Novartis Pharmaceuticals Corporation; Pfizer Inc.; Piramal Imaging; Servier; Takeda Pharmaceutical Company; and Transition Therapeutics. The Canadian Institutes of Health Research is providing funds to support ADNI clinical sites in Canada. Private sector contributions are facilitated by the Foundation for the National Institutes of Health (www.fnih.org). The grantee organization is the Northern California Institute for Research and Education, and the study is coordinated by the Alzheimer’s Therapeutic Research Institute at the University of Southern California. ADNI data are disseminated by the Laboratory for Neuro Imaging at the University of Southern California.

\section*{Declarations of interest statement}

 Frank Jessen reports fees for advice and presentations (2020-2023): AC immune, Biogen, Danone/Nutricia, Eisai, Grifols, Janssen, Lilly, Roche (unrelated to this study). The other authors do not declare any conflicts of interest.

\bibliographystyle{abbrv}

\bibliography{bibliography}

\clearpage

\newpage

\appendix

\section*{Appendix}

\setlength{\FrameRule}{0pt}
\setcounter{table}{0}
\setcounter{figure}{0}
\renewcommand{\thesubsection}{\Alph{subsection}}
\renewcommand\theequation{A.\arabic{equation}}
\renewcommand\thetable{A\arabic{table}}
\renewcommand\thefigure{A\arabic{figure}}

\subsection{Anisotropic Laplace level sets}\label{sec:albo}

The Laplace-Beltrami operator (LBO) $\Delta$ has been used widely for mesh/signal processing \cite{reuter:smi09,levy2006laplace,wetzler2013laplace} and shape analysis (initiated by our work on "ShapeDNA" \cite{reuter:cad06}, with applications in neuroimaging \cite{Wachinger-brainprint15}). The anisotropic Laplace-Beltrami operator ($\alpha$LBO) is an extension that models a directional dependency of the underlying diffusion process:
\begin{equation}
\Delta_A f = div( A (\nabla f) )
\end{equation} 
with the divergence ($div$), the gradient ($\nabla$) and a $2 \times 2$ matrix $A$ acting on the tangent vectors (with the identity representing the isotropic case). Inspired by \cite{andreux2014} who incorporate (extrinsic) surface curvature via the $\alpha$LBO into shape segmentation, we model anisotropy in a similar way, yet for solving the generalized eigenvalue problem. 

The first eigenfunction with smallest (non-zero) eigenvalue of the regular LBO (with Neumann boundary condition in case of a boundary) provides the smoothest embedding of a geometric shape onto the real line as it minimizes the Dirichlet energy. For eigenfunctions $f_i$ the Dirichlet energy resolves to the eigenvalue $\lambda_i$ :
\begin{equation}
\begin{aligned} 
E [ f_i ] :&= \int_\Omega \parallel \nabla f_i \parallel^2 d\sigma = -  \int_\Omega  f_i  \Delta f_i  d\sigma \\ &=  \lambda_i \int_\Omega  f_i  f_i  d\sigma = \lambda_i .
\end{aligned}
\end{equation}

In the isotropic case on a cylinder shell (with longer circumference than heights, similar to our hippocampal shapes) the first LBO eigenspace with Neumann boundary condition is two dimensional and the corresponding orthonormal basis are $\sin$ and $\cos$ (90 rotated). This basis can be rotated arbitrarily around the cylinder shell.
Using a curvature dependent anisotropic LBO, we can encourage the zero level sets of the first eigenfunction to settle on high (negative) curvature regions. This can be achieved by preferring or inhibiting diffusion along specific curvature directions.
We set the coefficients of the diagonal anisotropy matrix (defined in the orthonormal basis $(v_m, v_M)$ of the min and max curvature directions) to: 
\begin{equation}
A_{\alpha} = diag \left( exp(-\alpha_0 \ \lvert \kappa_M \rvert), exp(-\alpha_1 \ \lvert \kappa_m \rvert) \right)
\end{equation}
where $\kappa_m$ and $\kappa_M$ are the min and max curvature values. The operator is thus isotropic in planar regions. The $\alpha$ values control the level of anisotropy separately for the regions where the absolute max or min curvature is large. We set $\alpha_0 = 0$, to obtain isotropic diffusion also in regions with large positive curvature. A large positive $\alpha_1$ will strongly encourage zero level sets to localize at the desired negative curvature regions. Intuitively, modifying the diffusion of the operator is similar to modifying the metric (e.g.\ shrinking distances in specific directions) and using the regular LBO. Here, specifically, one can think of pinching the opposite boundaries of the cylinder shell towards each other, shrinking the height at high negative curvature areas. Due to the embedding theorem, zero level sets, where gradients are largest, will localize in those areas to minimize the Dirichlet energy. 

The algorithm to compute curvature (min, max curvature and directions) can be found, e.g., in \cite{alliez:acm03}. The implementation of the (anisotropic) Laplace operator is based on our FEM implementation \cite{reuter:cad06,reuter:smi09}. The linear FEM approach can be directly extended to tetrahedra meshes. Doing so will provide the same stiffness matrix as described for the mesh Laplace operator in e.g.\ \cite{wang2017towards}, however, with a different mass matrix. Our FEM mass matrix has the same sparsity pattern as the stiffness matrix and correctly represents the inner product of piecewise linear functions on the tetrahedral mesh. Similar to the 2D triangle case, the diagonal mesh Laplace mass matrix is a simplified version of the FEM mass matrix, obtained by lumping (summing) all elements to the diagonal. Another advantage of the FEM approach is its straight-forward extension to higher-order approximations with improved convergence properties (see e.g.\ \cite{strang2008}). We make our implementation of the discrete differential geometry operators for triangles and tetrahedral meshes, together with the FEM solvers, freely available as the Laplace Python (LaPy) library at \url{https://github.com/Deep-MI/LaPy} (also as pip and conda packages). LaPy provides highly efficient vectorized algorithms, making heavy use of sparse matrices and sparse FEM solvers. Our code for the hippocampal shape analysis (available at \url{https://github.com/Deep-MI/Hipsta} upon publication) requires LaPy as a dependency. 

\subsection{Curvature-based spatial alignment procedure}\label{sec:curv-align}

Here we provide a brief description of the curvature-based spatial alignment procedure that is used to correct for potential shifts of the coordinate system. The alignment procedure is taken from the field of functional data analysis, which deals with the analysis of signals (e.g., curves, surfaces) that vary over a continuum (often space or time). Across cases, such data is often characterized by amplitude variations as well as shifts in phase, which is also what we observe for the curvature profiles along the medial/lateral dimension. While both sources of variability can be interesting on their own, their joint presence confounds the interpretation of the signal. This can, however, be alleviated by an alignment, or registration, of the individual curves. The registration algorithm employs functional principal components analysis, which represents the signal in terms of a limited set of basis functions. In our scenario, this results in the estimation of a curvature template, which represents the dominant mode of variation across cases, as a first step. Step two is the estimation of a smooth warping function that maps the observed signals to the template. These two steps are repeated in an alternating manner until convergence of the algorithm.

\subsection{Supplementary experiments}\label{sec:increase}

\begin{figure}[!hbt]
\begin{framed}
\centering
\includegraphics[width=\textwidth]{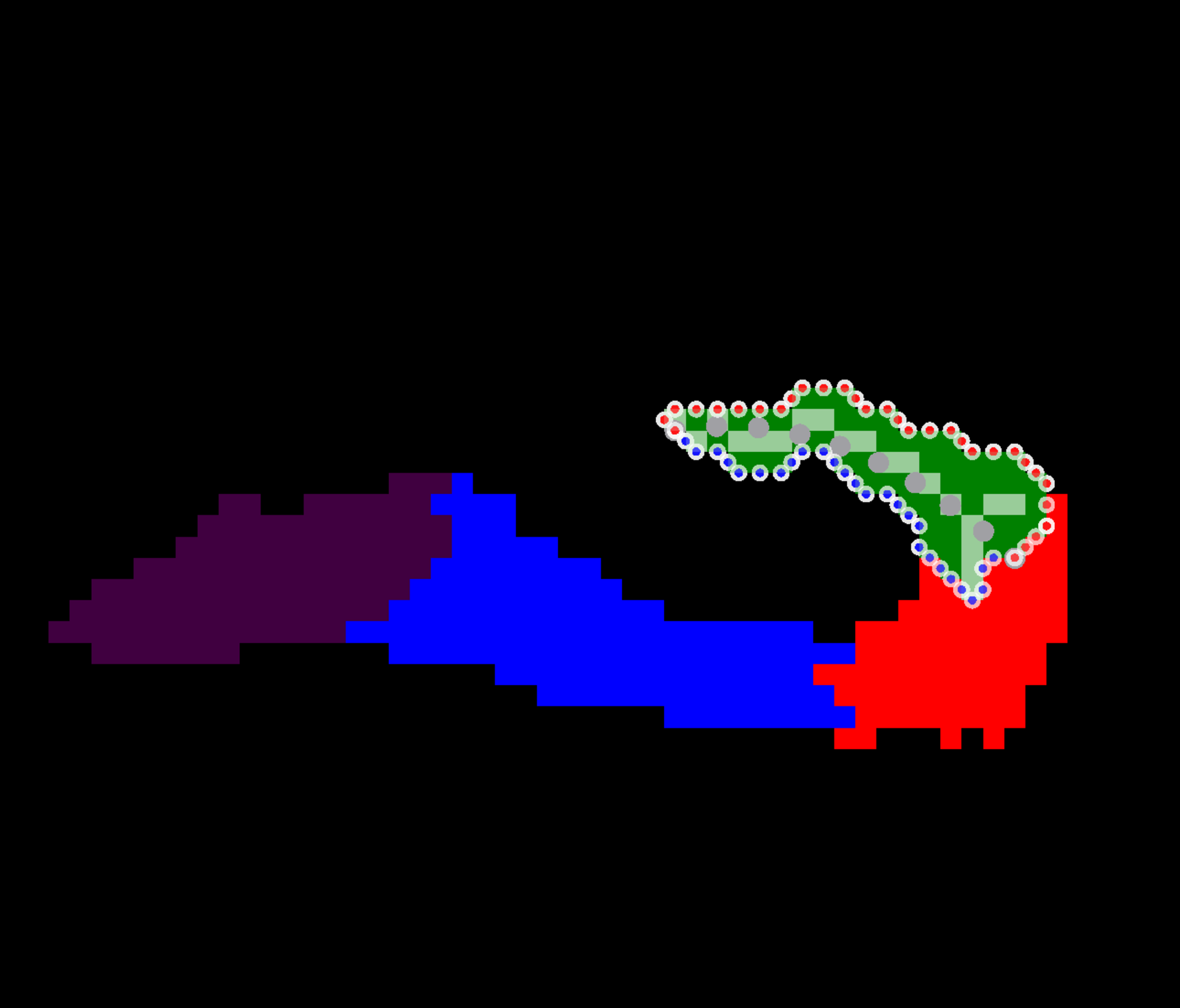}
\captionsetup{width=\textwidth}
\caption{Surrogate algorithm for image-based thickness estimation. Shown are hippocampal subfield labels from a FreeSurfer segmentation (purple: presubiculum, blue: subiculum, red: CA1, green: CA2/3) in a representative coronal slice. For the CA2/3 label, also its medial axis is shown (light green). Red and blue dots indicate outer and inner points on the contour line around the CA2/3 region. Grey dots indicate 10 equidistant points on a polynomial curve fitted to the medial axis. Thickness is measured between the red and blue points (except for those on the border between CA1 and CA2) that are closest to each grey dot.}
\label{fig:a-1}
\end{framed}
\end{figure}

We conducted an additional experiment to gain more insight into the unexpected thickness increases in the CA2/3 regions that were observed across analyses (Section~\ref{sec:replication}) and to rule out the possibility that these effects are introduced as an artifact of the proposed thickness estimation method. For this purpose, we devised a simplified thickness estimation algorithm that solely operates in image space (on a single slice) and does not depend on a mesh model of the hippocampus (Figure~\ref{fig:a-1}). Specifically, we extract the label for the CA2/3 region in a representative coronal slice and identify its medial axis. Next, we fit a polynomial of degree 2 to the medial axis and divide it into 30 segments of equal length (proportional sampling). To test if the distance of the sampling points has an influence, in a second analysis we use a varying number of segments with identical length (equidistant sampling). In both cases, for each segment, we find the two closest points on the inner and outer contours around the CA2/3 region (the points on the border between CA2 and CA1 are excluded). The distances between each pair of closest contour points are taken as crude image-based thickness estimates. 

\begin{figure}[!hbt]
\centering
\begin{framed}
\includegraphics[width=\textwidth]{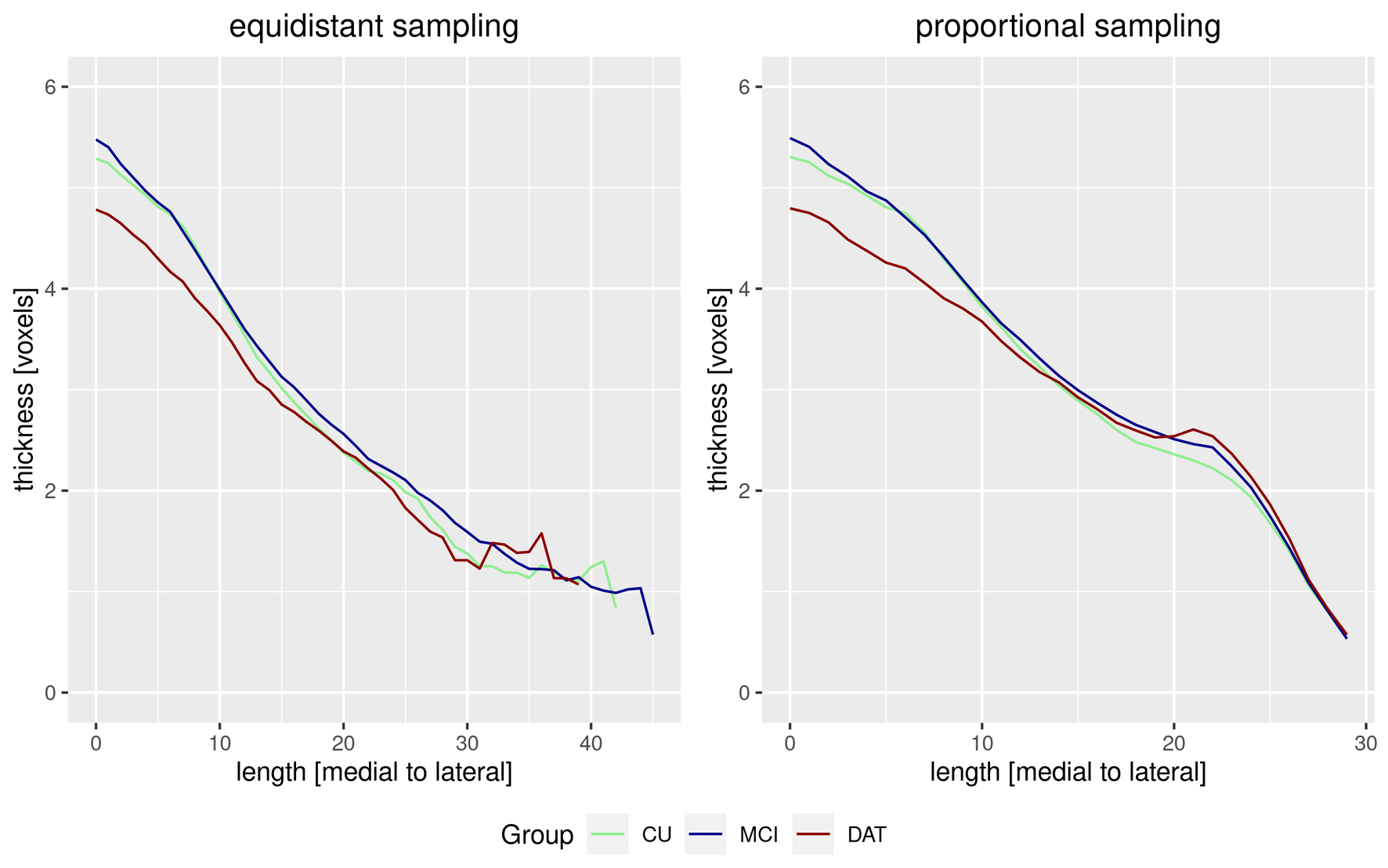}
\captionsetup{width=\textwidth}
\caption{Image-based thickness estimates (mean and 95\% confidence interval) for the three diagnostic groups in a FreeSurfer segmentation of the left hippocampus in the ADNI sample, also indicating subtle thickening in the DAT group in lateral regions (i.e., CA3).}
\label{fig:a-2}
\end{framed}
\end{figure}

Figure~\ref{fig:a-2} shows the simplified hippocampal thickness estimates for the two processing variants: for equidistant sampling, we find that there are less sample points on the medial axis in the DAT group than in the MCI or control groups, reflecting the reductions of length in the medial/lateral dimension as observed in our previous analyses of geometrical length measures across the hippocampal body (Section~\ref{sec:geometry}). For proportional sampling, all groups have the same number of sample points by definition (albeit with different spacing). Both processing variants also show thickness differences between groups: consistent with our previous analyses (Section~\ref{sec:replication}), reductions in thickness are primarily observed towards the medial end of the region under consideration, i.e.\ at the border between CA2 and CA1. For some parts of the CA2/3 region, in particular its tip, the analyses also show thickness increases in the DAT group as compared to the MCI group and the group of cognitively unimpaired controls, similar to our previous observation. Importantly, this is the case for both the proportional sampling and, although to a lesser extent, the equidistant sampling. These results suggest that a) the unexpected thickness increases are observed independently of the proposed, geometry-based thickness estimation algorithm, and b) that they are not, or at least not exclusively, due to differences in the medial/lateral extent of the hippocampus, and potentially ensuing differences in spatial sampling across anatomical subregions of the hippocampus.

\subsection{Supplementary analyses}

\subsubsection{Details and examples for QC procedures}\label{sec:qc}

A-priori anatomical QC by means of visual inspection and classification ("good", "fair", and "poor") of the segmentations has been performed for the ASHS segmentations in the DELCODE study, which is our main analysis sample (in fact, partly for this reason). In total, there were 379 (81\%) good, 32 (6.8\%) fair, and 57 (12.2\%) poor cases. A break-down of these numbers by diagnostic group showed that there were more lower-quality segmentations with disease progression: CU 2.5\%, MCI 19.5\% and DAT 23.1\% poor quality cases, which would be candidates for exclusion in an applied study. We, however, retain these cases in our analysis, as we are not primarily interested in generating new insights into AD pathology in this methods-oriented study and poor segmentation quality is not always directly connected to failure of the geometric method. 

For the other analyses and datasets, where no full QC was performed, we conducted a closer inspection of failure cases for both the main and the auxiliary analyses. This encompasses (A) failure analyses of the segmentation algorithms ("segmentation errors"), and (B) an analysis of cases where the thickness algorithm did not complete successfully ("geometry errors"). In total, we have observed failure rates of 5.8\% in the DELCODE/ASHS, 5.6\% in the DELCODE/FreeSurfer, 16.1\% in the ADNI/ASHS, and 5.3\% in the ADNI/FreeSurfer analyses. In the following, we report, unless noted otherwise, details on segmentation (A) and geometrical (B) errors for the ADNI/ASHS analysis. We have added examples for all of the following errors to the Appendix. We have observed similar errors also in the DELCODE/ASHS, DELCODE/FreeSurfer, and ADNI/FreeSurfer analyses. However, since the overall failure rates were much lower in these analyses, we did not do a detailed quantitative comparison as for the more problematic ADNI/ASHS analysis. 

\begin{figure}[h]
\centering
\begin{framed}
\begin{subfigure}[b]{\textwidth}
\includegraphics[width=\textwidth]{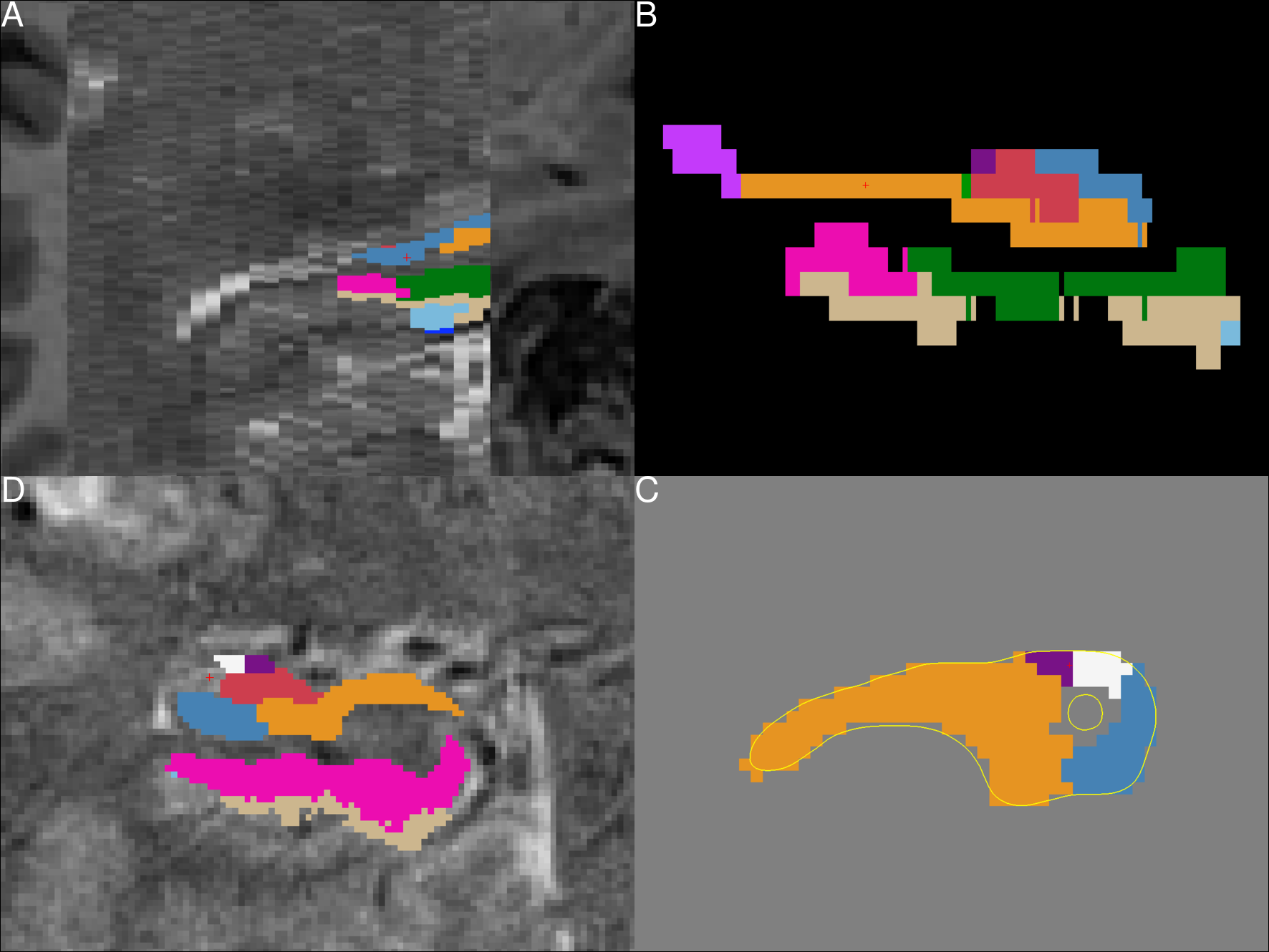}
\end{subfigure}
\captionsetup{width=\textwidth}
\caption{Examples of segmentation errors. A: insufficient coverage of the hippocampus by the high-resolution T2 scan. B: incorrect voxel dimensions (long edges in the inferior-superior direction instead of the anterior-posterior direction. C: Gap between CA1 (blue) and and CA2 (white). D: Connection between segmentations of the Subiculum (orange) and CA3 (purple).}
\label{fig:a-3}
\end{framed}
\end{figure}

(A) A major cause of failures are segmentation errors as shown in Figure \ref{fig:a-3}. Absolute segmentation failures (i.e., no output produced) were observed only rarely (only 90/1154 cases for ADNI/ASHS, none for the other analyses). Low-quality segmentations were observed more frequently than fundamental failures: Similar to the DELCODE/ASHS visual inspection results above, we observe such a poor segmentation quality in 12.5\% of the ADNI/ASHS cases, that no reasonable results could be expected from the application of our algorithm. 

\begin{figure}[h]
\centering
\begin{framed}
\begin{subfigure}[b]{\textwidth}
\includegraphics[width=\textwidth]{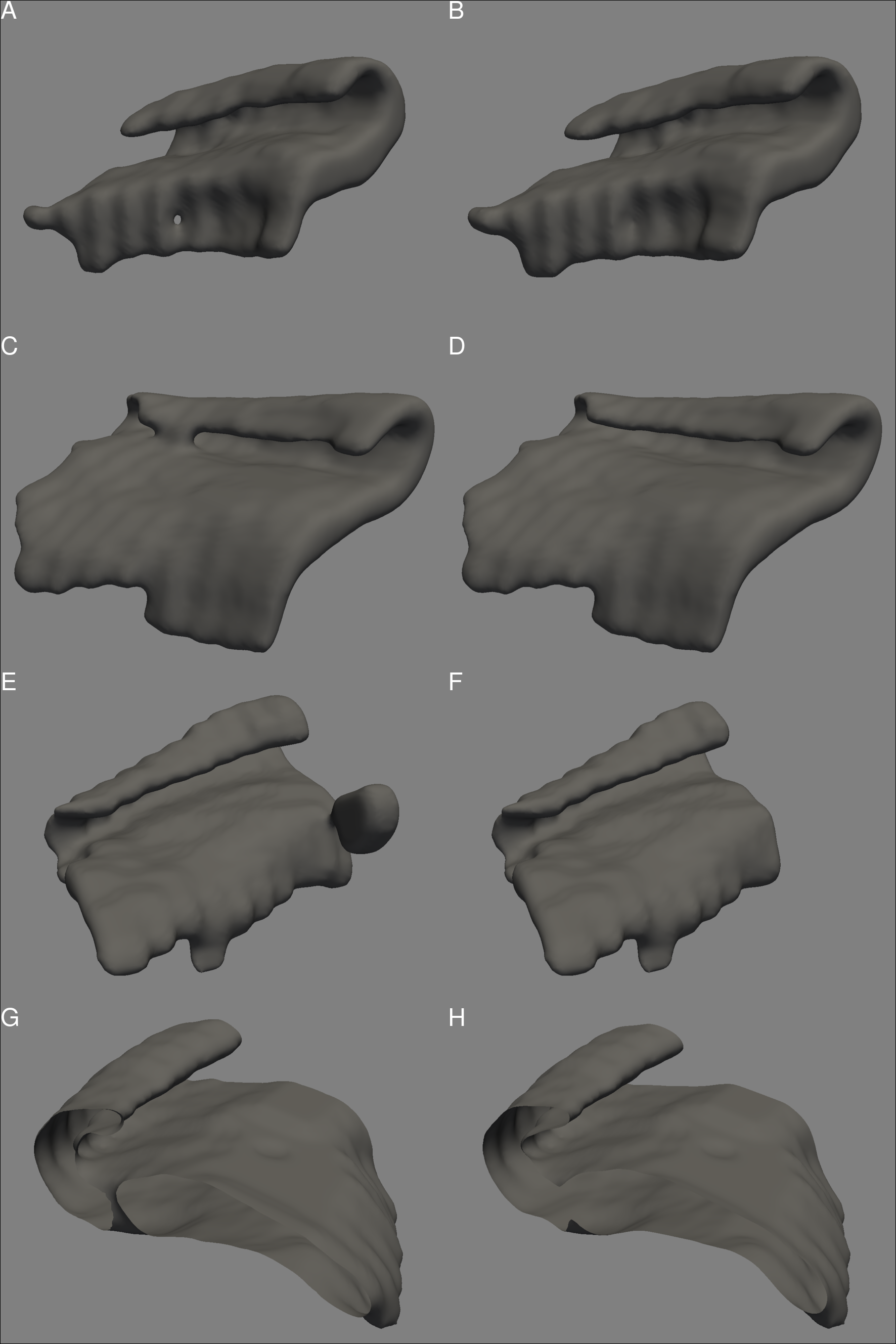}
\end{subfigure}
\captionsetup{width=\textwidth}
\caption{Examples of geometrical errors. A+B: holes, C+D: bridges, E+F: protrusions, G+H: boundary issues. Images on the left (A, C, E, G) are prior to correction, images on the right (B, D, F, H) after correction.}
\label{fig:a-4}
\end{framed}
\end{figure}

(B) Besides segmentation errors, the thickness estimation algorithm can fail or produce irregular output also because of errors that occur at a later stage ("geometrical errors", Figure~\ref{fig:a-4}). Here, we observed three classes of errors that prevent a successful completion of the algorithm: a) mesh errors (holes, bridges, protrusions): holes can be present in the mesh due to undersegmentation (missing voxels). Bridges can be present due to oversegmentation (e.g., anatomically implausible connections between CA3 and CA1 or the subiculum). Protrusions represent shape irregularities that can be due to either under- or oversegmenation; b) boundary errors: these are failures to correctly identify the boundaries of the mesh, either towards the head or tail, or at the medial and lateral ends. The latter could lead to improper placement of the mid-surface, which might not reach deep into each and every bend of the mesh. For noisy meshes this smooth LBO solution, however, could even have a regularizing effect; c) random errors make up the final class of geometrical errors with an unclear cause. In the ADNI/ASHS analysis, mesh errors were observed in 5.9\% of the cases, boundary errors were observed in 4.4\% of the cases, and random errors in 5.7\% of the cases.

To prevent and mitigate processing errors, we have implemented a set of measures in our algorithm to support the user. Specifically, we provide QC plots of the hippocampal surface and automatic identification of cases with holes, bridges, and protrusions, as well as boundary and mid-surface placement issues. If such issues are present, a specific error message is issued that describes the problem and ways to mitigate it (such as choosing different processing parameters).

\FloatBarrier

\subsubsection{Comparison of aligned vs.\ non-aligned thickness estimates}\label{sec:align}

\begin{figure}[h]
\centering
\begin{framed}
\begin{subfigure}[b]{\textwidth}
\includegraphics[width=\textwidth]{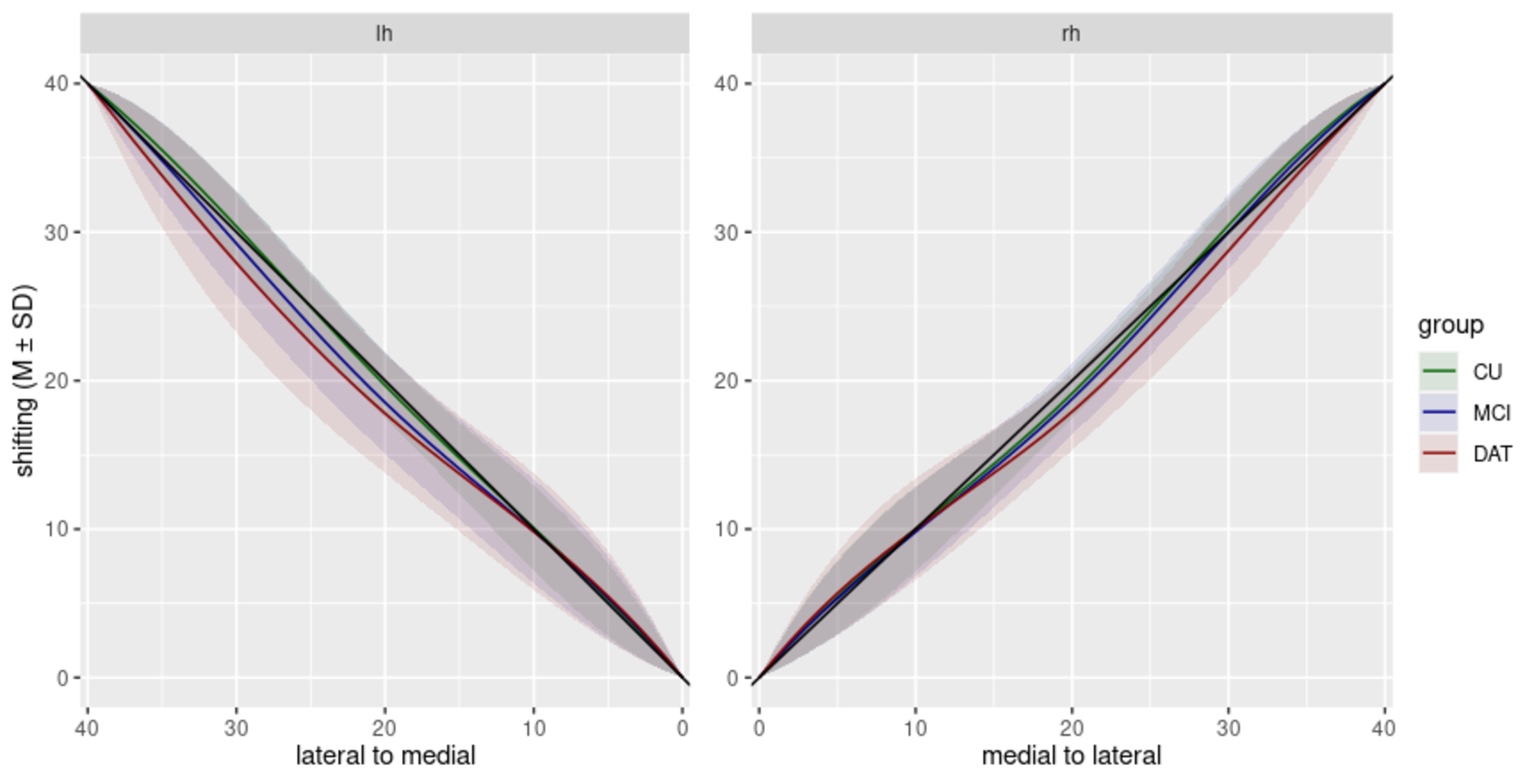}
\end{subfigure}
\captionsetup{width=\textwidth}
\caption{Descriptive analysis of shifting parameters for the curvature-based alignment. The $x$ axis shows original coordinates, and the $y$ axis shows the shifted positions of these coordinates. For reference, a straight line with zero intercept and unit slope indicates positions in case of no shifting.}
\label{fig:a-5}
\end{framed}
\end{figure}

We conducted an additional analysis to evaluate the impact of curvature alignment on the hippocampal thickness results. Figure \ref{fig:a-5} shows, for both hemispheres and the three diagnostic groups, the amount of shifting that is applied to individual $x$ coordinates. Specifically, the figure shows onto which new coordinate the original values are mapped. A comparison with a line through the origin and unit slope (depicted in black) shows that shifting is most pronounced in the middle, i.e.\ around $x=20$. For the largest part of the medial-lateral dimension, the shifting is in the medial direction, although a reversal can be observed at the medial end of the hippocampus. The amount of shifting is, on average, no more than 3 coordinates. The least amount of shifting is applied to the CU group, followed by the MCI and DAT groups. These observations are comparable in the left and right hemisphere.

\begin{figure}[h]
\centering
\begin{framed}
\includegraphics[width=\textwidth]{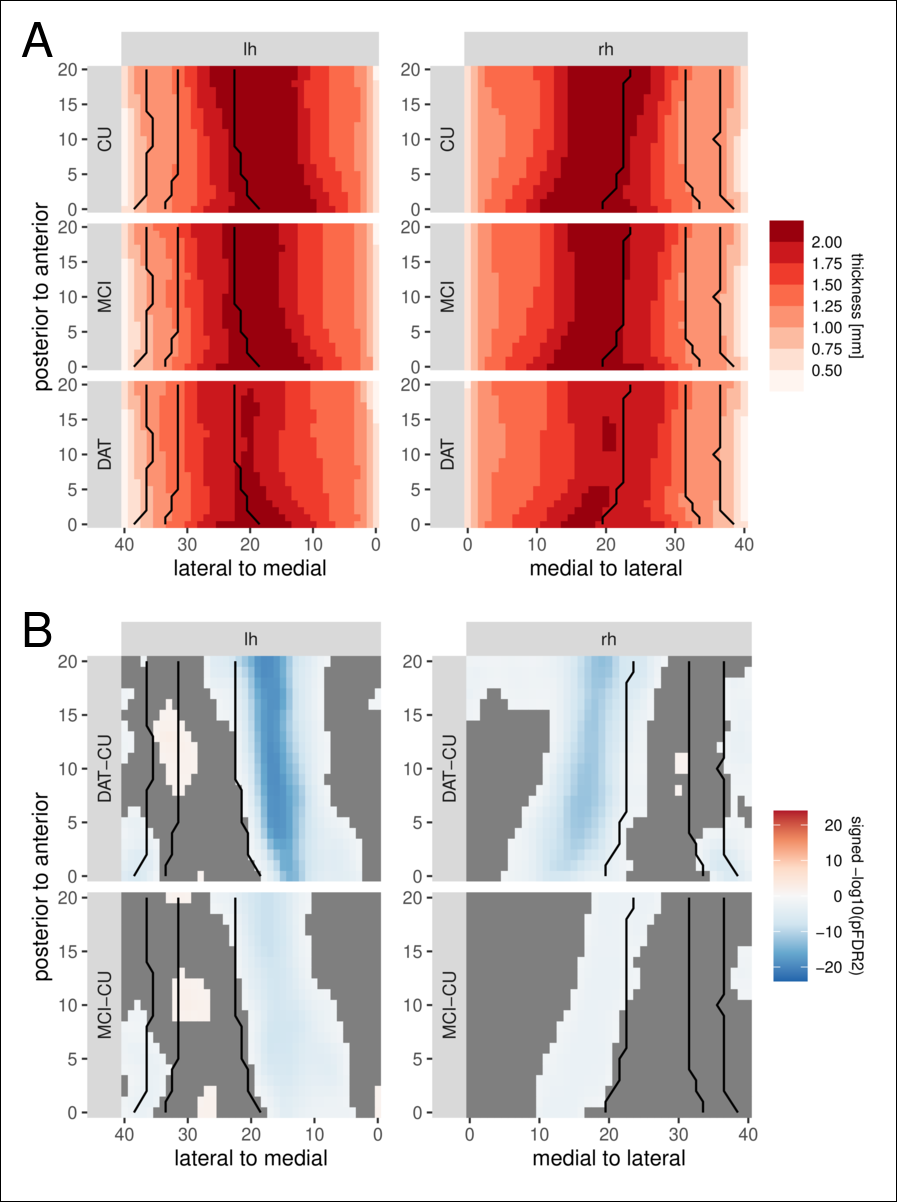}
\captionsetup{width=\textwidth}
\caption{Thickness estimates and evaluation of group differences for non-aligned data, based on the ASHS segmentation in the DELCODE sample. Top: mean localized hippocampal thickness estimates for the left and right hemisphere in the DAT, MCI, and control groups. Bottom: statistical evaluation of thickness differences between the DAT, MCI, and controls groups. Orange/blue colors indicate regions that are significant after FDR2-correction for multiple comparisons. Black lines indicate subfield boundaries between CA2/3, CA1, subiculum, and presubiculum, from lateral to medial.}
\label{fig:a-6}
\end{framed}
\end{figure}

Figure \ref{fig:a-6} shows thickness estimates and the statistical evaluation of group differences for \textit{non-aligned} data, and is intended to be compared with Figure \ref{fig:r-3}, which shows \textit{aligned} data. The overall appearance of the two figures is similar, both with regard to the spatial localization of thickness estimates and the magnitude of differences between, and both generally support the interpretation of more pronounced thickness differences for the $DAT$ vs.\ $CU$ comparison than for the $MCI$ vs.\ $CU$ comparison. The effect of shifting is relatively subtle; one notable difference between the aligned and non-aligned results is that for non-aligned data, apparent thickness increases in the DAT and MCI groups are observed in the subiculum and CA1 regions. We hypothesize that these are an artefact of shrinkage in the medial-lateral dimension in the DAT and MCI groups, which is not accounted for in the non-aligned data.

\FloatBarrier

\subsubsection{Incremental explanatory value of geometry-based measures}\label{sec:incr}

We evaluated whether geometrical measures (circumference, interior/exterior ratio, surface area, and shape index) provide incremental explanatory value for the classification of the MCI and DAT groups vs.\ cognitively unimpaired controls. As in the main part of the manuscript, we evaluate if the addition of geometrical measures improves the performance (area under the curve, AUC) of a logistic regression model that contains hippocampal volume as the only other predictor. Statistical significance of improvements is assessed using the likelihood ratio test. Table~\ref{tab-a-class-geom} shows that in contrast to hippocampal thickness, none of the geometrical measures improves the classification beyond hippocampal volume.

\begin{table}[ht]
\centering
\begin{subtable}{0.475\textwidth}
\centering
\begingroup\footnotesize
\begin{tabularx}{\textwidth}{L{1.1}R{0.9}C{1.3}R{0.7}}
\hline
& {\textbf{volume}} & {\textbf{volume \& circumference}} & {\textbf{LR test}} \\
\hline
DAT vs.\ CU & $0.90\pm 0.04$ & $0.90\pm 0.04$ & $0.300$ \\ 
MCI vs.\ CU & $0.73\pm 0.06$ & $0.73\pm 0.06$ & $0.084$ \\ 
\hline
& {\textbf{volume}} & {\textbf{volume \& int./ext.ratio}} & {\textbf{LR test}} \\ 
\hline
DAT vs.\ CU & $0.90\pm 0.04$ & $0.90\pm 0.04$ & $0.567$ \\ 
MCI vs.\ CU & $0.73\pm 0.06$ & $0.73\pm 0.06$ & $0.721$ \\ 
\hline
& {\textbf{volume}} & {\textbf{volume \& surface area}} & {\textbf{LR test}} \\ 
\hline
DAT vs.\ CU & $0.88\pm 0.05$ & $0.88\pm 0.05$ & $0.972$ \\ 
MCI vs.\ CU & $0.73\pm 0.06$ & $0.73\pm 0.06$ & $0.602$ \\ 
\hline
& {\textbf{volume}} & {\textbf{volume \& shape index}} & {\textbf{LR test}} \\ 
\hline
DAT vs.\ CU & $0.88\pm 0.05$ & $0.88\pm 0.05$ & $0.453$ \\ 
MCI vs.\ CU & $0.73\pm 0.06$ & $0.73\pm 0.06$ & $0.271$ \\ 
\hline
\end{tabularx}
\endgroup
\end{subtable}
\captionsetup{width=0.475\textwidth}
\caption{Classification performance (AUC$\pm$CI) and $p$-values of the likelihood ratio (LR) test for hippocampal volume and geometrical measures.}
\label{tab-a-class-geom}
\end{table}

\subsubsection{\textit{HippUnfold} results in the DELCODE dataset}\label{sec:hu-delcode-full}

\begin{figure}[!hb]
\centering
\begin{framed}
\includegraphics[height=0.8\textheight]{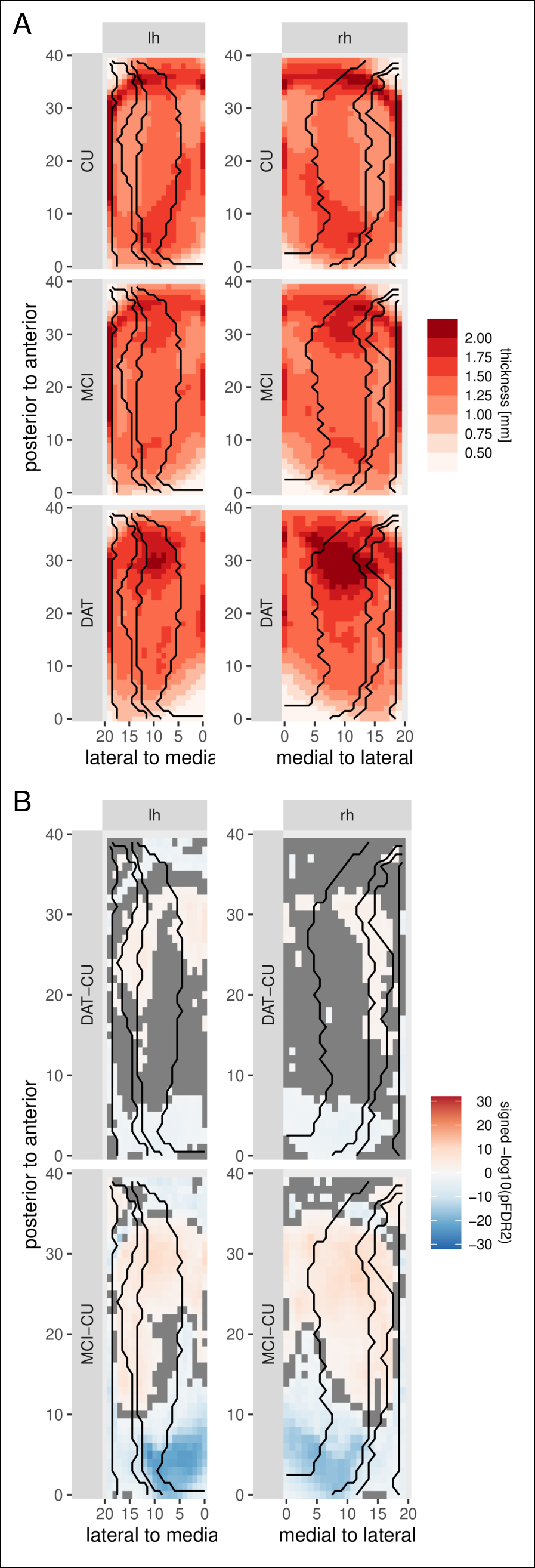}
\captionsetup{width=0.75\textwidth}
\caption{Evaluation of group differences in hippocampal thickness, based on \textit{HippUnfold} algorithm in the DELCODE sample. Top: mean localized hippocampal thickness estimates for the left and right hemisphere in the DAT, MCI, and control groups. Bottom: statistical evaluation of thickness differences between the DAT, MCI, and controls groups. Orange/blue colors indicate regions that are significant after FDR2-correction for multiple comparisons. Black lines indicate subfield boundaries between CA4, CA3, CA2, CA1, and subiculum, from lateral to medial.}
\label{fig:a-7}
\end{framed}
\end{figure}

Here we present a supplement to the analysis of the \textit{HippUnfold} results in the DELCODE dataset. While Section \ref{sec:replication} showed a cropped version for better comparison with the proposed algorithm, Figure \ref{fig:a-7} shows the non-cropped, full extent of the unfolded hippocampus. In addition to the effects in the hippocampal body, we now observe thickness reductions in the head in the DAT group as compared to the CU group, and an extension of a posterior cluster of thickness reductions in the DAT and MCI groups towards the tail.

\subsubsection{\textit{HippUnfold} results in the ADNI dataset}\label{sec:hu-adni-full}

\begin{figure}[!hb]
\centering
\begin{framed}
\includegraphics[height=0.8\textheight]{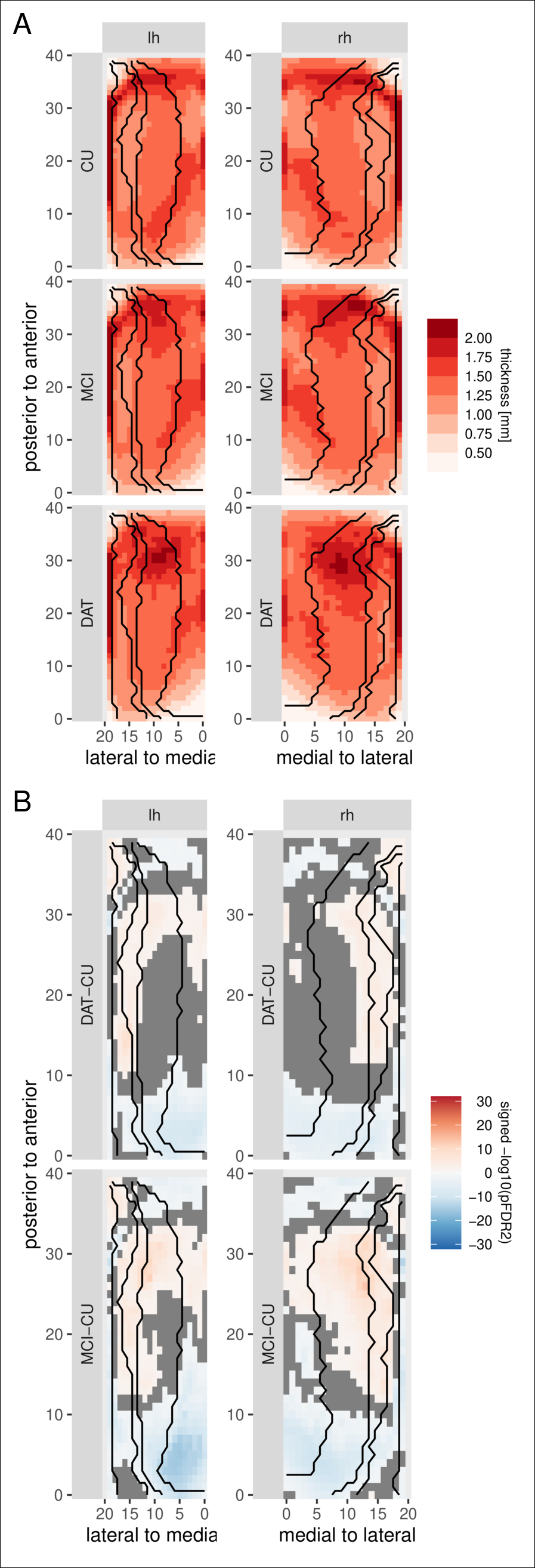}
\captionsetup{width=0.75\textwidth}
\caption{Evaluation of group differences in hippocampal thickness, based on \textit{HippUnfold} algorithm in the ADNI sample. Top: mean localized hippocampal thickness estimates for the left and right hemisphere in the DAT, MCI, and control groups. Bottom: statistical evaluation of thickness differences between the DAT, MCI, and controls groups. Orange/blue colors indicate regions that are significant after FDR2-correction for multiple comparisons. Black lines indicate subfield boundaries between CA4, CA3, CA2, CA1, and subiculum, from lateral to medial.}
\label{fig:a-8}
\end{framed}
\end{figure}

We conducted an additional analysis to evaluate whether or not the \textit{HippUnfold} results generalize across datasets. Figure~\ref{fig:a-8} shows the results of an application of the \textit{HippUnfold} algorithm in the ADNI data. We observe that thickness is highest in the most lateral regions in all groups. The statistical evaluation shows that thickness decreases in the DAT group, and to a lesser in the MCI group, are primarily present in posterior regions of the hippocampus. Further, thickness increases appear to be present in more anterior regions covering CA1, CA2 and the subiculum. Altogether, the results closely resemble the results that were obtained for the DELCODE dataset.

\raggedbottom % workaround to achieve better vertical spacing on this page

\end{document}